\documentclass[11pt]{article}

\usepackage{amsmath}
\usepackage{amsthm}
\usepackage{amssymb}
\usepackage{color}
\usepackage{graphicx,psfrag,epsf}
\usepackage{enumerate}
\usepackage{natbib}
\usepackage{subfigure}
\usepackage{authblk}
\usepackage{parskip}
\usepackage{xcolor}
\usepackage{multirow}
\usepackage[linesnumbered,ruled,vlined]{algorithm2e}

\newcommand{\blind}{0}

\usepackage{helvet}

\theoremstyle{plain}
\newtheorem{theorem}{Theorem}[section]

\theoremstyle{remark}

\newcommand{\M}[1]{\boldsymbol{#1}}  
\newcommand{\V}[1]{\boldsymbol{#1}}  

\begin{document}

\if0\blind
{
	{\title{\bf Sequential Adaptive Design for Jump Regression Estimation
			\author{Chiwoo Park, Peihua Qiu, Jennifer Carpena-Núñez, Rahul Rao, Michael Susner, and Benji Maruyama}
		}
	}
	\maketitle
} \fi
\if1\blind
{
	\bigskip
	\bigskip
	\bigskip
	\title{\bf Sequential Adaptive Design for Jump Regression Estimation}
	\author{}
	\date{}
	\maketitle
} \fi

\begin{abstract}
Selecting input variables or design points for statistical models has been of great interest in adaptive design and active learning. Motivated by two scientific examples, this paper presents a strategy of selecting the design points for a regression model when the underlying regression function is discontinuous. The first example we undertook was for the purpose of accelerating imaging speed in a high resolution material imaging; the second was use of sequential design for the purpose of mapping a chemical phase diagram. In both examples, the underlying regression functions have discontinuities, so many of the existing design optimization approaches cannot be applied because they mostly assume a continuous regression function. Although some existing adaptive design strategies developed from treed regression models can handle the discontinuities, the Bayesian approaches come with computationally expensive Markov Chain Monte Carlo techniques for posterior inferences and subsequent design point selections, which is not appropriate for the first motivating example that requires computation at least faster than the original imaging speed. In addition, the treed models are based on the domain partitioning that are inefficient when the discontinuities occurs over complex sub-domain boundaries. We propose a simple and effective adaptive design strategy for a regression analysis with discontinuities: some statistical properties with a fixed design will be presented first, and then these properties will be used to propose a new criterion of selecting the design points for the regression analysis. Sequential design with the new criterion will be presented with comprehensive simulated examples, and its application to the two motivating examples will be presented.
\end{abstract}

\noindent%
\vfill

\section{Introduction}
Regression analysis is a powerful statistical tool for estimating a regression function that relates explanatory variables to a response variable. In a typical regression analysis, the design points are assumed to be given in advance. When the design points can be selected during a data collection process, optimizing the selection is referred to as optimal design \citep{chernoff1972sequential}, active learning \citep{cohn1996active}, or adaptive sensing \citep{arias2013fundamental, malloy2014near}. This paper aims to address the problem of selecting the design points for a regression model particularly when the underlying regression function is piecewise continuous, motivated by two scientific applications. 

\begin{figure}[t]
	\includegraphics[width=0.9\textwidth]{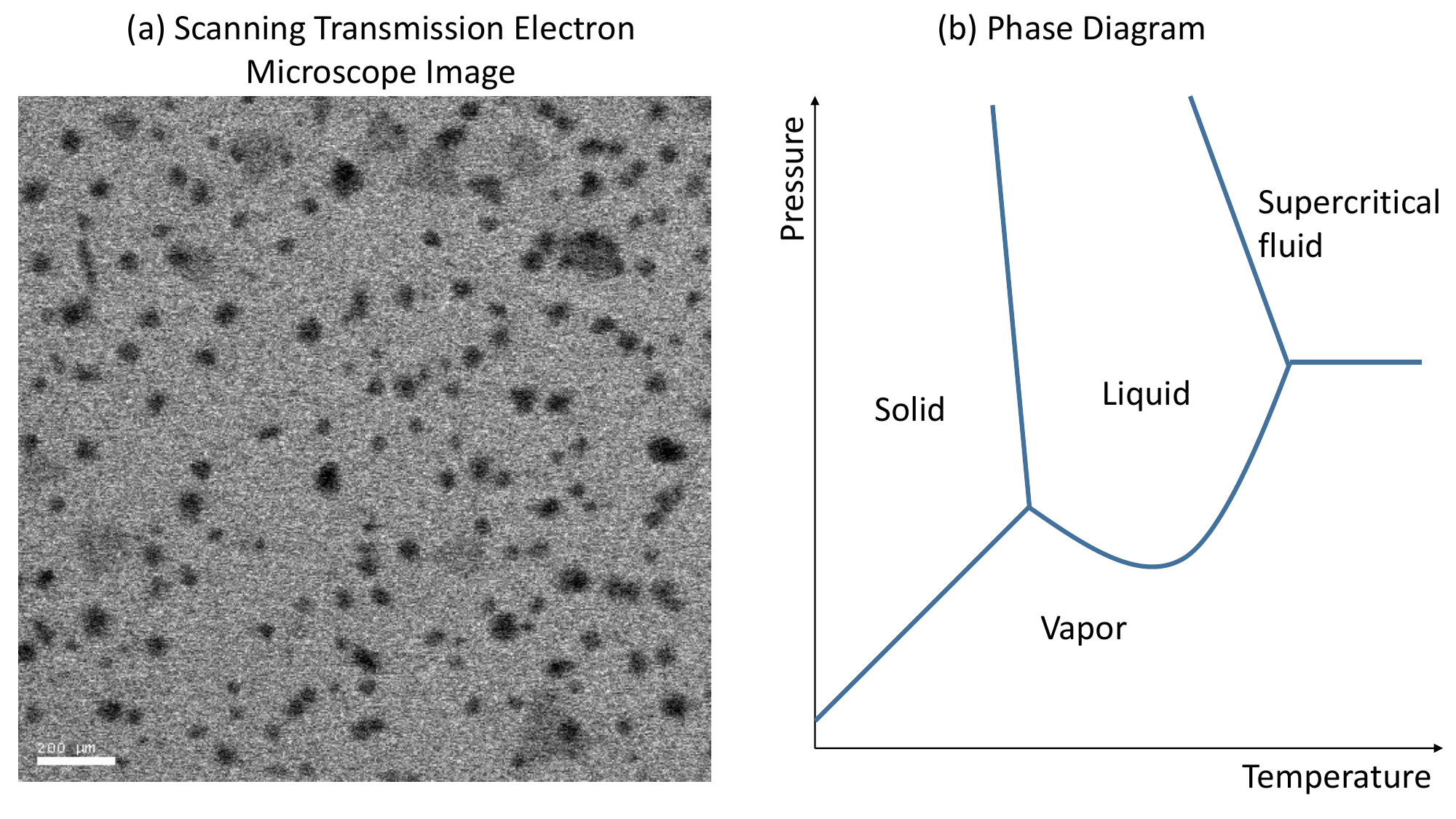}
	\caption{Two motivating examples of design optimization in jump regression analysis.} \label{fig:example}
\end{figure}

The first motivating application is the material imaging with scanning transmission electron microscopy (STEM). The STEM technique is a very important material characterization tool to image the microstructures of a material specimen at a sub-angstrom spatial resolution. It uses a focused beam of electrons to probe a material specimen, and the intensity of the beam interacting with the specimen is measured for every focus location. This sequential imaging process will create a rastered image of the material specimen as shown in Fig. \ref{fig:example}-(a), where a pixel corresponds to one focus location over the raster, and the corresponding intensity becomes the pixel intensity.  The radius of the focused beam can be below an angstrom or $10^{-10}$ meter, which allows a specimen to be imaged at a very fine spatial resolution; however, this level of detail is also the major reason for a slow imaging speed. The poor temporal resolution limits the STEM to be applied for studying material samples dynamically evolving in time. The time of the beam to stay at a pixel location is a pixel dwell time. The total imaging time is equivalent to the pixel dwell time multiplied by the total number of pixel locations. To maintain the high signal-to-noise ratio, the pixel dwell time cannot be reduced too much. The existing approaches to accelerate the imaging speed are based on a partial scan to scan a material specimen only at selected pixel locations. The partial set is randomly selected from an uniform distribution over the raster locations in the existing approaches \citep{stevens2015applying}, which can cause loss in spatial resolution. Optimizing the pixel locations in the partial set is highly desirable to mitigate the information loss. In this example, selecting the pixel locations can be formulated as optimizing the design points in a regression analysis, where the intensity surface of the image is regarded as a function of a 2D pixel coordinate. As shown in Fig. \ref{fig:example}-(a), a typical material image has smooth variation in intensity both in the image background and in the objects being imaged, the latter of which is called image foreground. However, there are significant jumps in intensity between the image foreground and background. Therefore, the underlying regression function in this example would be a piecewise continuous function in 2D. 

The second application is to optimize the design of experiments for effectively exploring a chemical phase diagram in chemistry. A phase diagram is a map that relates different experimental conditions to physical states of materials. The physical state suddenly jumps from one state to another around the experimental conditions where phase transitions occur, as illustrated in Fig. \ref{fig:example}-(b). Typically the elucidation of a phase diagram requires a large number of experiments to be performed to probe possible physical states that may exist in the experimental phase-space.  Optimizing the design of experiments is thus essential for an effective probing process. In particular, we are interested in carbon nanotube growth experiments to study the chemical conditions required for good nanotube growth. The chemical conditions include the reaction temperature and a relative ratio of two chemical ingredients, a reducing agent and an oxidant. The total nanotube growth changes abruptly around the boundary condition in the relative ratio for a given temperature. Therefore, the total nanotube growth is a discontinuous function of the relative ratio and temperature. 

In both motivating applications described above, the underlying regression functions are piecewise continuous in low-dimensional design spaces. In the existing studies of sequential design and active learning, the underlying regression function is always assumed to be a continuous function, and many design selection approaches have been developed under that assumption as we will review in Section \ref{sec:review}. There are some existing adaptive design approaches that are based on tree-based regression models, which can handle the discontiuities, e.g., decision tree \citep{malloy2014near, goetz2018active}, dynamic trees \citep{taddy2011dynamic}, and Bayesian treed regression models \citep{gramacy2009adaptive}. Those Bayesian approaches inherently come with computationally expensive MCMC samplings, which are computationally too slow to be feasible for our first motivating application. Please recall that the major goal of adaptive design in the first application is to reduce the scan time. If an adaptive design strategy was applied, then the overall scan time would be the physical scan time plus the computation time used for design selection. If the computation time surpasses the full image scan time (about 10 seconds for scanning an entire 500 $\times$ 500 image), then the adaptive design approach would be useless. Our proposed approach is based on jump regression modeling, which is a simple and computationally efficient regression analysis techinique for estimating a piecewise continuous regression function \citep{qiu2005image}, and we proposed adaptive design strategy runs faster than the full image scan.  

The remainder of this paper is organized as follows. In Section \ref{sec:review}, we review the existing research in active learning and sequential design, and emphasize the needs for a new adaptive design for our two motivating examples. Section \ref{sec:method} describes an approach of jump regression analysis for estimating a discontinuous regression function and discusses its statistical properties in cases with a fixed design. This approach is then used for developing a novel sequential adaptive design strategy for regression analysis with discontinuities. Section \ref{sec:experiment} presents numerical studies with a number of simulated examples. Section \ref{sec:app1} presents the application of the proposed approach to the first motivating application, and Section \ref{sec:app2} illustrates the application to the second motivating application. Finally, Section \ref{sec:conclude} concludes the article with some summary statements.

\section{Related Work} \label{sec:review}

The design optimization problem has been studied in the experimental design and active learning literatures. Some existing approaches are briefly reviewed in this section. 

In experimental design, the relationship between experimental factors and an experimental outcome is often described by a parametric regression model. Optimal experimental design exploits such a relationship for selecting a design of experiments that would result in a better parameter estimation \citep{sacks1989design}. Most literature focuses on batch or open-loop designs that choose the design of all experiments concurrently, so the experimental designs are not affected by experimental outcomes. Sequential experimental design allows experiments to be conducted sequentially, exploiting past experimental outcomes to guide the design of future experiments. In many existing approaches, the sequential design was considered as a problem of augmenting an initial fixed design by a sequentially chosen set of design points. For a given parametric model, the data from the initial design points are used to estimate the model parameters, the next batch of design points are selected so as to optimize a design criterion, and the design criterion is typically chosen to be the same as those used in the open-loop design strategies, including the D-optimality and I-optimality. \cite{chaudhuri1993nonlinear} used the D-optimality that maximized the determinant of the fisher information matrix of the estimated parameters from the past experiments. \cite{sinha2002robust} used the minimization of the integrated mean square error as a criterion, corresponding to the I-optimality criterion in the open-loop design. \cite{dror2008sequential} also extended the D-optimality criterion under the Bayesian framework to better accommodate the sequential design for a small size of design points. 

The sequential design for nonparametric regression models has also been developed. \cite{zhao2012sequential} discussed the sequential design problem in the context of kernel regression, based on the mean integrated square error criterion. \cite{bull2013spatially} studied a similar problem in cases with a univariate nonparametric regression model that was estimated by the wavelet decomposition approach. Gaussian process regression models and the related design problems have been studied for spatial data analysis, but that remains in finding an optimal open-loop design using the maximum entropy criterion \citep{zhu2006spatial, zimmerman2006optimal}. 

In active learning, selecting the design points was studied for a broader set of nonparametric regression models, such as the Gaussian process regression models \citep{krause2008near, singh2009efficient, hoang2014nonmyopic} and the kernel-based regression models \citep{paisley2010active}. These existing approaches have been developed mainly for regression modeling with a continuous regression function. 

There are a few existing adaptive design strategy for the tree-structure regression models, which can accommodate discontinuities in regression analysis. In the treed models, the input domain is partitioned into subdomains, and a simple regression model is posed for each sub-domain. \citet{malloy2014near, goetz2018active} studied adaptive learning strategies for a piecewise constant regression function using decision trees. \cite{bull2013spatially} also discussed an active learning strategy for spatially inhomogeneous regression functions including piecewise constant functions and functions with sharp bumps, but it was limited to regression with a single explanatory variable.  \citet{gramacy2009adaptive} discussed a treed Gaussian process model and the corresponding sequential design strategies, involving a treed-partitioning and GP leaves. Its posterior inference involved computationally inefficient reversible-jump methods for MCMC or higher-dimensional particles for sequential inference. \citet{taddy2011dynamic} proposed a dynamic tree model for a simple leave model, constant or linear leave model and an adaptive design strategy for the model. It is more computationally efficient than the treed GP model but the former is more restrictive. Another potential limitation of the approaches is that they partition the regression domain recursively along axis-aligned directions, which can be ineffective for discontinuities occuring over complex boundaries, creating many little partitions around the complex boundaries. 

Another potentially related approach is the adaptive design for estimating the contour of the underlying regression function $m(x)$ \citep{ranjan2008sequential}, which sequentially selects the design points to estimate the contour of the subregion $m(x) < c$ for some constant c. When the discontinuities in $m(x)$ occur at a constant level $c$, estimating the contour at c would create a partition of the regression domain around the discontinuities so allow to put several regression models on the partitions. The major limitation in applying this approach for the two motivating applications is that the level $c$ should be known as priori, and its estimation is not straightforward, because the discontinuities do not always occur at a single level $c$ (e.g., material images on uneven background such as MG5 and MG6 in Fig. \ref{fig:mgdata} in the first motivating application).

The jump regression analysis \citep{qiu2005image} provides a simpler modeling approach for a broader class of piecewise continuous functions, and it is more computationally efficient. The proposed modeling approach meets the time constraints and the model adequacy of the two motivating applications. 

\section{Method} \label{sec:method}
Let $\mathcal{X}$ denote a closed subset of $\mathbb{R}^p$ that represents a design space in a regression modeling problem. We consider a general jump regression model that aims to estimate a nonparametric regression function $m: \mathcal{X} \rightarrow \mathbb{R}$ from its noisy observations,
\begin{equation}
Y_i := m(\V{x}_i) + \epsilon_i, 
\end{equation}
where $\{Y_i; i=1,\ldots, n\}$ are noisy observations of the response variable $Y$ at the design points $\{\V{x}_i \in \mathcal{X}; i =1,\ldots, n\}$, and $\{\epsilon_i; i=1,\ldots, n\}$ are random errors with mean zero and variance $\sigma^2$. The underlying regression function $m$ is further assumed piecewise continuous, such that there exists a partition $\{\mathcal{A}_b; b = 1,\ldots,B\}$ of the design space $\mathcal{X}$ satisfying 
\begin{enumerate}
	\item[(a)] Each $\mathcal{A}_b$ is a simple connected (nonempty) subset of $\mathcal{X}$, 
	\item[(b)] $\cup_{b=1}^B \mathcal{A}_b = \mathcal{X}$, and $\mathcal{A}_b \cap \mathcal{A}_{b'} = \emptyset$, for any $b \neq b'$,
	\item[(c)] The function $m(\V{x})$ has the expression
	\begin{equation*}
	m(\V{x}) = \sum_{b = 1}^B g_b(\V{x}) I_{\mathcal{A}_b}(\V{x}), \qquad \mbox{ for } \V{x} \in \mathcal{X}.
	\end{equation*} 
	where $g_b(\V{x}) \in \mathcal{C}^2(\mathcal{X})$ is a smooth function, for each $b$. Thus, the regression function $m(\V{x})$ is continuous in $\mathcal{A}_b \backslash \partial \mathcal{A}_b$, where $\partial \mathcal{A}_b$ is the boundary set of $\mathcal{A}_b$, and 
	it has jumps over $\mathcal{B} := \cup_{b=1}^B \partial \mathcal{A}_b$. For any $\V{x}^* \in \mathcal{B}$, there exists $b$ and $b'$ such that $\V{x}^* \in \partial \mathcal{A}_b \cap \partial \mathcal{A}_{b'}$ and 
	\begin{equation*}
	  \lim_{\V{x} \rightarrow \V{x}^*, \V{x} \in \mathcal{A}_b} g_b(\V{x}) \neq
          \lim_{\V{x} \rightarrow \V{x}^*, \V{x} \in \mathcal{A}_{b'}} g_{b'}(\V{x}).
	\end{equation*}
	The boundary set $\mathcal{B}$ is referred to as the jump location curves (JLCs) of $m$ in the literature \citep{qiu1998discontinuous}. 
	\item[(d)] The boundary is smooth, so a tangent line exists almost everywhere. A point in the boundary set $\mathcal{B}$ is called non-singular when there exists a unique tangent line at the point. Otherwise, it is called singular. We denote a collection of all singular boundary points by $\mathcal{S}$. For $\V{x}^* \in \mathcal{B} \backslash \mathcal{S}$, there exist a unique pair of $b$ and $b'$ such that $\V{x}^* \in \partial \mathcal{A}_b \cap \mathcal{A}_{b'}$. Otherwise, its tangent line would not be unique. 
	\item[(e)] The jump size between $\mathcal{A}_b$ and $\mathcal{A}_{b'}$ at $\V{x}^*$ is defined as
	\begin{equation*}
	\delta_{b, b'}(\V{x}^*) = \lim_{\V{x} \rightarrow \V{x}^*, \V{x} \in \mathcal{A}_b} g_b(\V{x}) - \lim_{\V{x} \rightarrow \V{x}^*, \V{x} \in \mathcal{A}_{b'}} g_{b'}(\V{x}), \V{x}^* \in \partial \mathcal{A}_b \cap \partial \mathcal{A}_{b'}.
	\end{equation*}
	It is assumed that $\delta_{b, b'}(\V{x}^*) \neq 0$ and they have the same sign, for any $\V{x}^* \in \partial \mathcal{A}_b \cap \partial \mathcal{A}_{b'}$. 
\end{enumerate}

Estimation of $m(\V{x})$ has been studied using two different approaches. By the first approach, the partition $\{\mathcal{A}_b; b=1,\ldots, B\}$ and the corresponding JLCs are estimated first, and then $m(\V{x})$ is estimated using the conventional local smoothing procedures (e.g., kernel smoothing methods) within each sub-region $\mathcal{A}_b$ \citep{qiu1997jump}. By the second approach, the regression function $m(\V{x})$ is estimated by one-sided kernel smoothing estimate, without explicit estimation of the JLCs \citep{qiu2009jump}. However, optimizing the selection of the design points in a jump regression model has not been studied in these papers. The current paper aims to develop a design selection strategy for jump regression analysis, primarily intended for the two scientific applications mentioned in the introduction, but it is general enough to be applied to other similar problems. To describe this design selection strategy, we first discuss how we estimate $m(\V{x})$ in a fixed design case in Section \ref{sec:fixed}, and then move on to the selection of design points in a sequential design setup in Section \ref{sec:seqdes}.

\subsection{Regression function estimation in a fixed design case} \label{sec:fixed}
Given observations $Y_1, \ldots, Y_n$ at the design points $\{\V{x}_1, \ldots, \V{x}_n\}$, we discuss nonparametric estimation of $m(\V{x})$, based on the one-sided local linear kernel smoothing approach \citep{qiu2009jump}. In this paper, we extend the approach with two modifications for our scientific applications. First, it is assumed that the design points are sparsely located in $\mathcal{X}$, and their locations are non-uniformly distributed over $\mathcal{X}$ as a result of optimizing the choice of design points in sequential design cases and other reasons. To accommodate such non-uniformly distributed design points, we use spatially varying kernel bandwidth, instead of a constant bandwidth used in \cite{qiu2009jump}. Second, we extend the method from 2D cases (i.e., $\V{x}$ has a dimension of 2) to cases with two or more dimensions. In all of our scientific applications, the dimension is two, but we hope the proposed approach can be more broadly applicable for other applications with more than two dimensions, although we still assume the dimension is quite small (e.g., three or four). 

The one-sided kernel smoothing approach does not require explicit estimation of $\{\mathcal{A}_b\}$, and it gives a pointwise estimate of the regression function $m(\V{x})$ directly with the jumps in $m(\V{x})$ be accommodated. For a given location $\V{x} \in \mathcal{X}$, consider its neighborhood with the bandwidth $h$: 
\begin{equation*}
\mathcal{N}(\V{x}) = \{\V{x}' \in \mathcal{X}: d(\V{x}', \V{x}) \le h \},
\end{equation*}
where $d(\cdot, \cdot)$ is the Euclidean distance. We seek a local estimate of $m(\V{x})$ using observed data in the neighborhood $\mathcal{N}(\V{x})$. In cases when the design points are uniformly distributed in $\mathcal{X}$, a global bandwidth parameter is typically used as a function of the sample size $n$. In this paper, we allow the design points $\{\V{x}_1, \ldots, \V{x}_n\}$ to be sampled from a non-uniform density $f(\V{x})$, due to the design selection procedure that will be discussed in the next section. To be more adaptive to the non-uniform density, we adopt spatially varying bandwidth parameters. Let $h_n(\V{x})$ denote the location-dependent bandwidth parameter, which is set to be the Euclidean distance from $\V{x}$ to its $k$th nearest neighbor (k-NN) in $\{\V{x}_1, \ldots, \V{x}_n\}$. The corresponding neighborhood of $\V{x}$ is defined to be 
\begin{equation*}
\mathcal{N}_n(\V{x}) := \{\V{x}' \in \mathcal{X}: d(\V{x}', \V{x}) \le h_n(\V{x}) \}.
\end{equation*}
Based on the existing literature on the k-NN density estimation \citep{wasserman2006all}, the k-NN bandwidth selection is asymptotically equivalent to selecting the bandwidth parameter to be inversely proportional to the density of the design points, i.e., 
\begin{equation} \label{eq:band}
h_n(\V{x}) \propto \left(\frac{1}{nf(\V{x})}\right)^{1/p}.
\end{equation}
Based on that asymptotic relation, $k$ should be chosen such that $k = o(n)$ and $k \rightarrow \infty$, as $n \rightarrow \infty$ \citep{mack1979multivariate}. Our choice in this paper is $k = \sqrt{n}$.

\begin{figure}[t]
	\includegraphics[width=\textwidth]{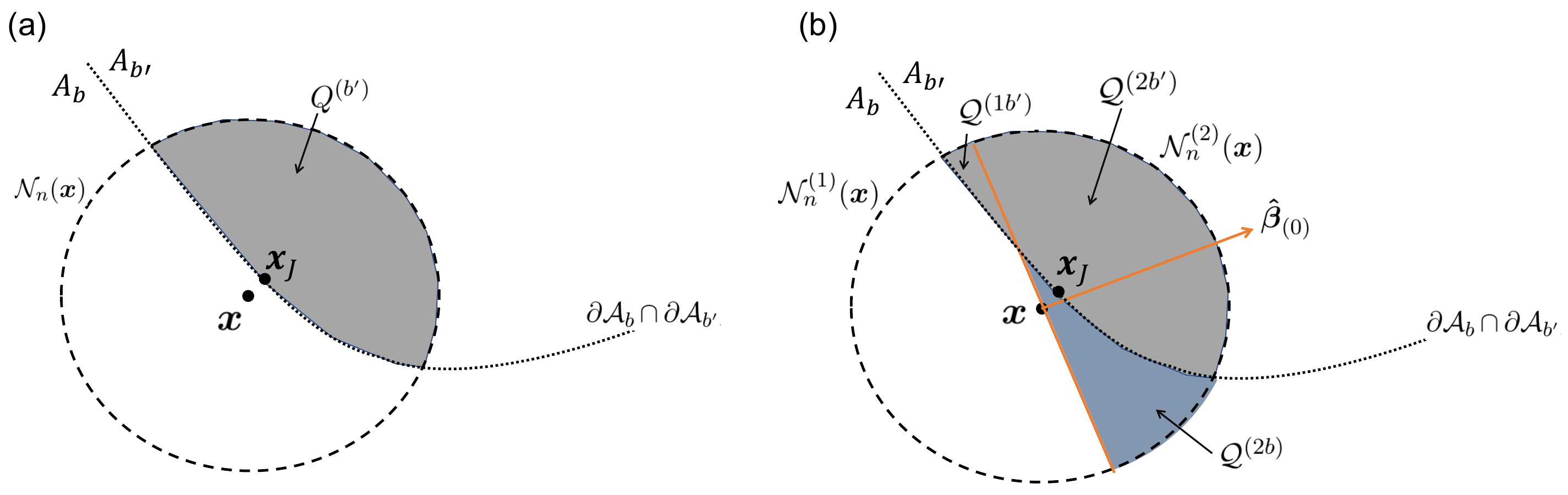}
	\caption{(a) Conventional local linear kernel estimate uses observations in a local neighborhood $\mathcal{N}_n(\V{x})$, (b) one-sided local linear kernel estimate uses observations in one of the two halves of $\mathcal{N}_n(\V{x})$.} \label{fig:method}
\end{figure}

For the conventional local linear kernel smoother, a local estimate of $m(\V{x})$, for any location $\V{x} \in \mathcal{X}$, is taken using available observations in the local neighborhood $\mathcal{N}_n(\V{x})$. Different from the conventional approach, the one-sided local linear kernel estimate is obtained using the observations in one of the two halves of $\mathcal{N}_n(\V{x})$. The split of $\mathcal{N}_n(\V{x})$ into two halves is made so that at least one of them is asymptotically on one side of the JLCs. To proceed, we first describe the conventional local linear kernel estimate and its error for estimating a jump regression function in order to motivate the needs for the one-sided estimate. Let $\hat{m}_{(0)}(\V{x})$ denote the conventional local linear estimate of $m(\V{x})$, which is the solution to $\alpha$ of the following optimization problem:
\begin{equation} \label{eq:llocal}
\min_{\alpha, \V{\beta}} \sum_{\V{x}_i \in \mathcal{N}_n(\V{x})} \left[Y_i - \alpha - \V{\beta}^T(\V{x}_i-\V{x}) \right]^2 K\left( \frac{\V{x}_i - \V{x}}{h_n(\V{x})}\right),
\end{equation}
where $K(\V{u})$ is an isotropic kernel function with a unit-circle support $\{\V{u} \in \mathbb{R}^p: \V{u}^T\V{u} \le 1\}$. The following theorem gives the asymptotic bias and variance of the estimate:
\begin{theorem} \label{thm:1}
	Assume that $g(\V{x}) \in C^2(\mathcal{X})$ has a bounded second-order derivative, the kernel $K$ is a Lipschitz-1 continuous and isotropic density function, and $h_n(\V{x})$ follows \eqref{eq:band}. For a given point $\V{x} \in \mathcal{A}_b$, if the projection of the point to the boundary set $\mathcal{B}$ is $\V{x}_J$ and it is non-singular, i.e., $\V{x}_J \in \mathcal{B} \backslash \mathcal{S}$, then there exists a unique pair of $b$ and $b'$ such that $\V{x}_J \in \partial \mathcal{A}_{b} \cap \partial \mathcal{A}_{b'}$, and 
	\begin{equation} \label{eq:bias_a}
	\begin{split}
	E[\hat{m}_{(0)}(\V{x})] - m(\V{x}) =& o_P\left( \frac{1}{n^{2/p} f(x)^{2/p}}\right) + (c_{J} + o_P(1))\int_{\mathcal{Q}^{(b')}} K(\V{u}) d\V{u},
	\end{split}
	\end{equation}
	and
	\begin{equation} \label{eq:var_a}
	Var[\hat{m}_{(0)}(\V{x})|\V{x}_1,\ldots,\V{x}_n] = \kappa_1 \sigma^2 (1+ o_P(1)),
	\end{equation}
	where $c_J = \delta_{b, b'}(\V{x}_J)$ is the jump magnitude at $\V{x}_J$, $\kappa_1$ is a constant depending on the kernel function, and $Q^{(b')}$ is the part of the kernel support that corresponds to $\mathcal{A}_{b'} \cap \mathcal{N}_n(\V{x})$.
\end{theorem}
\noindent The proof of the theorem is provided in the Online Supplementary Material. From \eqref{eq:bias_a} and \eqref{eq:var_a}, the variance of the estimate is asymptotically a constant. The bias is significantly affected by $d(\V{x}, \V{x}_J)$, the distance of the test point $\V{x}$ to the nearest jump location curve. Please note that if $d(\V{x}, \V{x}_J) \ge h_n(\V{x})$, i.e., the test point is far away from the jump location curve, then $\mathcal{A}_{b'} \cap \mathcal{N}_n(\V{x}) = \emptyset$ and consequently $\mathcal{Q}^{(b')}$ is an empty set. In such a case, the bias is simply $o_P\left( \frac{1}{n^{2/p} f(x)^{2/p}}\right)$, which is same as the bias of the conventional local linear kernel estimate in a continuous region. However, when the distance goes below $h_n(\V{x})$, $\mathcal{Q}^{(b')}$ is non-empty, as illustrated in Fig. \ref{fig:method}-(a). In such cases, the additional bias, $c_{J} \int_{\mathcal{Q}^{(b')}} K(\V{u}) d\V{u}$, is generated. The additional bias is bounded above by 
\begin{equation} \label{eq:bnd_bias}
c_{J} \int_{\mathcal{Q}^{(b')}} K(\V{u}) d\V{u} \le c_{J}K\left(\frac{\V{x} - \V{x}_J}{h_n(\V{x})}\right) \mathcal{L}(\mathcal{Q}^{(b')}),
\end{equation}
where $\mathcal{L}(\cdot)$ is the Lebesgue measure, and $\mathcal{L}(\mathcal{Q}^{(b')})$ is
\begin{equation*}
O_p\left( \max\left\{0,1-\left(\frac{d(\V{x}_J, \V{x})}{h_n(\V{x})}\right)^p\right\} \right).
\end{equation*}
Therefore, this part of the bias increases as $d(\V{x}, \V{x}_J) / h_n(\V{x})$ decreases, i.e., the test point approaches to the JLC. 

To mitigate the bias increment, the local neighborhood $\mathcal{N}_n(\V{x})$ is halved into $\mathcal{N}_n^{(1)}(\V{x})$ and $\mathcal{N}_n^{(2)}(\V{x})$, by a plane passing through $\V{x}$ and being perpendicular to $\V{\hat{\beta}}_{(0)}$, as illustrated in Fig. \ref{fig:method}-(b), where $\V{\hat{\beta}}_{(0)}$ is the solution to $\V{\beta}$ in the conventional local linear estimation \eqref{eq:llocal}. According to Corollary 1 in \cite{qiu2009jump}, $\V{\hat{\beta}}_{(0)}$ is approximately perpendicular to the tangent plane of the jump location curve at $\V{x}_J$ with some approximation error. Therefore, the cutting plane is approximately in parallel to the tangent plane of the jump location curve, and either one of the two halves would be approximately on one side of the jump location curve. For example, in Fig. \ref{fig:method}-(b), the test point $\V{x}$ is in $\mathcal{A}_b$, and $\mathcal{N}_n^{(1)}(\V{x})$ mostly belongs to $\mathcal{A}_b$ except for its small portion that corresponds to $\mathcal{Q}^{(1b')}$ in the figure.     


In each one-sided neighborhood $\mathcal{N}_n^{(l)}(\V{x})$, for $l=1,2$, we take the one-sided local linear kernel estimate of $m$, denoted as $\hat{m}_{(l)}(\V{x})$, to be the solution of $\alpha$ to the following optimization problem,
\begin{equation} 
(\hat{m}_{(l)}(\V{x}), \hat{\V{\beta}}_{(l)}(\V{x})) = \arg\min_{\alpha, \V{\beta}} \sum_{\V{x}_i \in \mathcal{N}^{(l)}_n(\V{x})} \left[Y_i - \alpha - \V{\beta}^T(\V{x}_i-\V{x})\right]^2 K\left( \frac{\V{x}_i - \V{x}}{h_n(\V{x})}\right).
\end{equation}
The final estimate of $m(\V{x})$ is chosen to be one of $\hat{m}_{(1)}(\V{x})$ and $\hat{m}_{(2)}(\V{x})$, and the choice depends on their estimation errors. The bias and variance of the two one-sided estimates are given in Theorem \ref{thm:2}. The proof of Theorem \ref{thm:2} is similar to that of Theorem \ref{thm:1}.
\begin{theorem} \label{thm:2}
	Under the same conditions stated in Theorem \ref{thm:1}, we have
	\begin{equation}
	\begin{split}
	E[\hat{m}_{(1)}(\V{x})] - m(\V{x}) =& o_P\left( \frac{1}{n^{2/p} f(x)^{2/p}}\right) \\
	&+ (2c_{J}+ o_P(1)) \int_{\mathcal{Q}^{(1b')}} K(\V{u}) d\V{u}, \\
	E[\hat{m}_{(2)}(\V{x})] - m(\V{x}) =& o_P\left( \frac{1}{n^{2/p} f(x)^{2/p}}\right) \\
	&+ (- 2c_{J}+o_P(1)) \int_{\mathcal{Q}^{(2b')}} K(\V{u}) d\V{u},
	\end{split}
	\end{equation}
	and
	\begin{equation}
	Var[\hat{m}_{(l)}(\V{x})|\V{x}_1,\ldots,\V{x}_n] = 2\kappa_1\sigma^2 (1+ o_P(1)),
	\end{equation}
	where $Q^{(lb')}$ is the part of the kernel support that corresponds to $\mathcal{A}_{b'} \cap \mathcal{N}^{(l)}_n(\V{x})$. 
\end{theorem}
\noindent By the above theorem, the variances of the two one-sided estimates are asymptotically the same. Therefore, the mean squared errors of the estimates are largely influenced by their respective bias terms. The major parts of the asymptotic biases are $2c_{J} \int_{\mathcal{Q}^{(1b')}} K(\V{u}) d\V{u}$ and $2c_{J} \int_{\mathcal{Q}^{(2b')}} K(\V{u}) d\V{u}$. Since $\mathcal{Q}^{(1b')} \cup \mathcal{Q}^{(2b')} = \mathcal{Q}^{(b')}$, the two terms can be written as
\begin{equation*}
\begin{split}
& 2c_{J} \int_{\mathcal{Q}^{(1b')}} K(\V{u}) d\V{u} = a_1 c_{J} \int_{\mathcal{Q}^{(b')}} K(\V{u}) d\V{u} \mbox{ and }\\
& 2c_{J} \int_{\mathcal{Q}^{(2b')}} K(\V{u}) d\V{u} = (2-a_1) c_{J} \int_{\mathcal{Q}^{(b')}} K(\V{u}) d\V{u},
\end{split}
\end{equation*}
for a constant $a_1 \in [0, 2]$. The smaller value of the two terms is bounded above by
\begin{equation} \label{eq:bnd_onebias}
\begin{split}
& 2 c_J \min\left\{\int_{\mathcal{Q}^{(1b')}} K(\V{u}) d\V{u}, \int_{\mathcal{Q}^{(2b')}} K(\V{u}) d\V{u} \right\} \\
& \quad \le c_{J}K\left(\frac{\V{x} - \V{x}_J}{h_n(\V{x})}\right) O_P\left( \max\left\{0,1-\left(\frac{d(\V{x}_J, \V{x})}{h_n(\V{x})}\right)^p\right\} \right) \min\{a_1, 2-a_1\}.
\end{split}
\end{equation}
The last term, $\min\{a_1, 2-a_1\}$, depends only on $\V{\hat{\beta}}_{(0)}$. When $\V{\hat{\beta}}_{(0)}$ is along the tangent plane at $\V{x}_J$, the value of $\min\{a_1, 2-a_1\}$ is approximately at its maximum, one. When the direction of $\V{\hat{\beta}}_{(0)}$ is perpendicular to the tangent plane, this value is zero. In the literature, it has been confirmed that $\V{\hat{\beta}}_{(0)}$ is asymptotically perpendicular to the tangent plane \citep{qiu2009jump}. Thus, $\min\{a_1, 2-a_1\}$ is approximately zero.

The bias terms cannot be numerically evaluated since $\mathcal{Q}^{(1b')}$ and $\mathcal{Q}^{(2b')}$ are unknown. To make a choice between $\hat{m}_{(1)}(\V{x})$ and $\hat{m}_{(2)}(\V{x})$, the following weighted residual 
mean errors are considered:
\begin{equation*}
err^{(l)}(\V{x}) = \frac{\sum_{\V{x}_i \in \mathcal{N}^{(l)}_n(\V{x})} \left[Y_i - \hat{m}_{(l)}(\V{x}) - \hat{\V{\beta}}_{(l)}(\V{x})^T (\V{x}_i - \V{x}) \right]^2  K\left( \frac{\V{x}_i - \V{x}}{h_n(\V{x})}\right)}{\sum_{\V{x}_i \in \mathcal{N}^{(l)}_n(\V{x})} K\left( \frac{\V{x}_i - \V{x}}{h_n(\V{x})}\right)}.
\end{equation*}
When $err^{(1)}(\V{x}) < err^{(2)}(\V{x})$, $\hat{m}_{(1)}(\V{x})$ is chosen; and $\hat{m}_{(2)}(\V{x})$ is chosen otherwise. 

\subsection{Proposed Method for Sequential Design Selection} \label{sec:seqdes}
In this section, we describe our proposed method for a sequential selection of design points, which takes the design points in multiple stages. The first stage serves as a seed stage, and the design points for the first stage are randomly sampled from an uniform distribution or can be selected by the Latin hypercube sampling (LHS). In all of our numerical examples, we used LHS. Each of the subsequence stages can be described as follows. Suppose that there are $n$ design points selected up to the previous stage, and we describe how $b$ additional design points are selected in the new stage. Let $f_1$ denote the unknown density of the $n$ design points from the previous stages, and let $f_{2|1}$ represent the sampling density used to draw the $b$ design points for the next stage. If $f$ was a `desirable' joint density of the $n$ design points and the $b$ additional design points, then the sampling density for the next stage's design points should be the conditional density conditioned on $f_1$, i.e.,
\begin{equation} \label{eq:joint}
f_{2|1}(\V{x}) = \frac{f(\V{x})}{f_1(\V{x})}.
\end{equation}
Intuitively, $f$ should be chosen to minimize the integrated square loss,
\begin{equation*}
\int_{\V{x} \in \mathcal{X}} E[m(\V{x}) - \hat{m}(\V{x})]^2 d\V{x}.
\end{equation*}
Note that the square loss $E[m(\V{x}) - \hat{m}(\V{x})]^2$ can be decomposed into the squared-bias and the variance of $\hat{m}(\V{x})$. Based on Theorem \ref{thm:2}, the variance of the jump regression estimate defined in Section \ref{sec:fixed} is approximately a constant, and the square loss is largely influenced by the squared-bias term. The bias can be as small as $o_P\left( \frac{1}{n^{2/p} f(x)^{2/p}}\right)$ when the test location is far away from the jump location curve in the sense that $d(\V{x},\V{x}_J) \ge h_n(\V{x})$. If $d(\V{x},\V{x}_J) < h_n(\V{x})$, there is an additional bias, $2 c_J \min\left\{\int_{\mathcal{Q}^{(1b')}} K(\V{u}) d\V{u}, \int_{\mathcal{Q}^{(2b')}} K(\V{u}) d\V{u} \right\}$. By the result \eqref{eq:bnd_onebias}, that part of the bias is bounded above by 
\begin{equation*}
c_{J}K\left(\frac{d(\V{x}_J, \V{x})}{h_n(\V{x})}\right) O_P\left( \max\left\{0,1-\left(\frac{d(\V{x}_J, \V{x})}{h_n(\V{x})}\right)^p\right\} \right) \min\{a_1, 2-a_1\},
\end{equation*}
which goes to zero as $\frac{d(\V{x}_J, \V{x})}{h_n\V{x}}$ increases or $d(\V{x}_J, \V{x})f(\V{x})$ increases. To balance off the bias over $\V{x}$ and minimize the integrated square loss function, the desirable sampling density should be 
\begin{equation*}
f(\V{x}) \propto \frac{1}{d(\V{x}_J, \V{x})}.
\end{equation*}
We hope that collectively the $n+b$ design points have higher densities at places near the jump location curve (i.e., places with small $d(\V{x}_J, \V{x})$). Certainly, we do not know where the jump location curves are located in practice, so we do not know the distance $d(\V{x}_J, \V{x})$. But, the distance can be roughly located using the observations of the regression function at the $n$ design points selected by the previous stages. It is easy to show that the following statistic increases as $d(\V{x}, \V{x}_J)$ decreases, 
\begin{equation} \label{eq:jump_stat}
[\hat{m}_{(1)}(\V{x}) - \hat{m}_{(2)}(\V{x})]^2,
\end{equation} 
so we use it as a jump detection statistic. Based on the jump detection statistic, we propose a desirable joint density $f$ to be
\begin{equation} \label{eq:fden}
f(\V{x}) = C\exp\left\{\alpha[\hat{m}_{(1)}(\V{x}) - \hat{m}_{(2)}(\V{x})]^2\right\}, \V{x} \in \mathcal{X},
\end{equation} 
where $C >0$ is a normalization constant, and the coefficient $\alpha$ controls the exploration vs exploitation trade-off. We chose $\alpha = 1/\sigma^2$, where $\sigma^2$ is the noise variance. With that choice, the quantity $\alpha(\hat{m}_1-\hat{m}_2)^2$ is approximately the square of the jump magnitude relative to the noise variance. For higher $\sigma^2$, this sampling function seeks more exploration, and for lower $\sigma^2$, more exploitation is sought. The noise standard deviation $\sigma$ is estimated using the median absolute deviation (MAD). 

Because $\mathcal{X}$ is bounded and the estimates $\hat{m}_{(l)}$ are bounded, $C$ is well defined. From \eqref{eq:joint}, the sampling density for the $b$ new design points should be
\begin{equation} \label{eq:conden}
f_{2|1}(\V{x}) = \frac{f(\V{x})}{f_1(\V{x})} \approx \frac{C\exp\left\{(\hat{m}_{(1)}(\V{x}) - \hat{m}_{(2)}(\V{x}))^2\right\}}{\frac{1}{n}\sum_{i=1}^{n} K\left(\frac{\V{x}-\V{x}_i}{h}\right)},
\end{equation}
where the approximation comes from the standard kernel density estimation of $f_1(\V{x})$, and $h$ is a non-spatial adaptive kernel bandwidth parameter that depends on the sample size $n$. Sampling from the complex density \eqref{eq:conden} can be performed by the Metropolis-Hasting Algorithm. For more computational feasibility, we can limit the sampling locations to ones sampled from the uniform distribution over the regression domain. For each of the possible sampling locations, we can compute $f_{2|1}(\V{x})$ up to a normalizing constant. The computed values are normalized so that the summation of all the computed values is equal to one. The normalized values will serve as the probability mass function (pmf) defined on a finite number of the possible sampling locations, and then $b$ i.i.d. samples will be taken from that pmf as the $b$ new design points for the next stage.


\section{Simulation Study: 2D and 3D Domains} \label{sec:experiment}
\begin{figure}[t]
	\includegraphics[width=\textwidth]{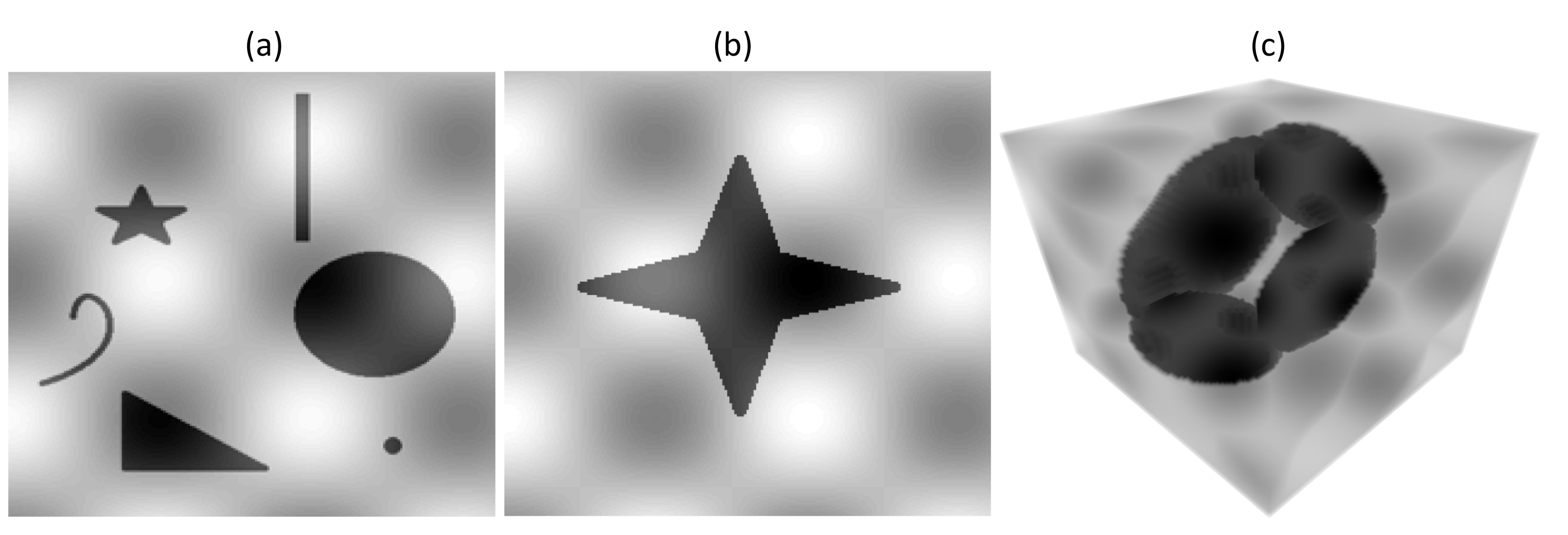}
	\caption{Three synthetic datasets. (a) 2d-others, (b) 2d-star, (c) 3d-donut} \label{fig:orig}
\end{figure}

For the initial validation of the proposed method, we performed a simulation study with three synthetic datasets. Fig. \ref{fig:orig} shows the underlying noise-free regression functions for the first two synthetic datasets defined on the 2D domain $[0, 200]^2$ and also shows the regression function for the third dataset defined on the 3D domain $[0, 50]^3$. The underlying noise-free regression functions are in the mixture form,
\begin{equation*}
m(\V{x}) = g_0(\V{x}) - 0.3 I_{\mathcal{A}_b}(\V{x}),
\end{equation*}
where $g_0$ is continuous functions on $\mathcal{X}$, and $\mathcal{A}_b \subset \mathcal{X}$ represents the subregion with a different intensity level. The regression function is continuous except at the boundary $\partial \mathcal{A}_b \subset \mathcal{X}$. For the first two datasets,
\begin{equation*}
g_0(\V{x}) = \sin\left(\frac{x_1}{20}\right) \times \cos\left(\frac{x_2}{20}\right),
\end{equation*}
where $x_1$ and $x_2$ are the first and second elements of the input vector $\V{x}$ respectively. For the last dataset, we used
\begin{equation*}
g_0(\V{x}) = \sin\left(\frac{x_1}{5}\right) \times \cos\left(\frac{x_2}{5}\right) \times \sin\left(\frac{x_3}{5}\right),
\end{equation*}
where $x_1$, $x_2$ and $x_3$ are the first, second and third elements of $\V{x}$ respectively. In Fig. \ref{fig:orig} and Fig. \ref{fig:orig3}, the set $\mathcal{A}_b$ is represented as the dark regions. We added i.i.d.\ Gaussian noise from $\mathcal{N}(0, \sigma^2)$ to $m(\V{x})$. 

\begin{figure}[ht]
	\includegraphics[width=\textwidth]{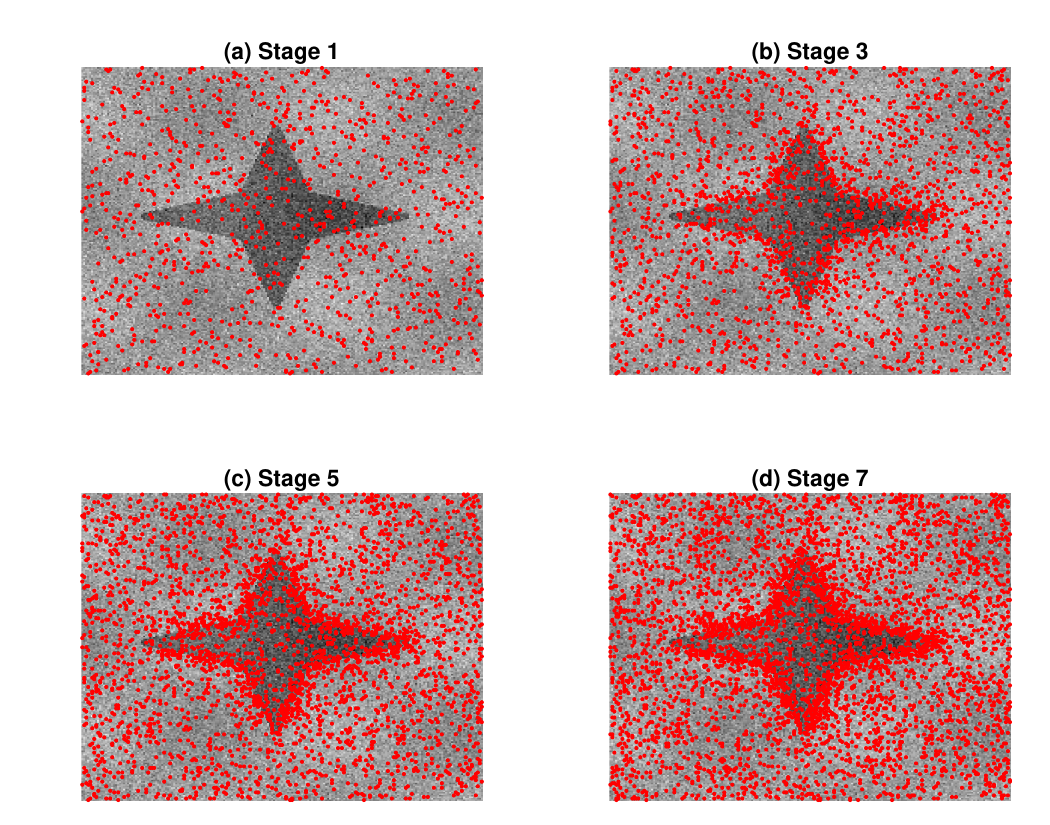}
	\caption{Illustration of selecting design points over stages. Each plot shows all the design points selected up to the specified stage.} \label{fig:orig8}
\end{figure}

For each dataset, $n$ design points in total are selected, using our sequential adaptive approach described in Section \ref{sec:seqdes}. We varied the number $n$ and the number of the design points taken for each stage, denoted by $b$. Since the domain sizes differ for the datasets, 200 x 200 for the first two datasets and 50 x 50 x 50 for the last dataset, the experimental settings are denoted in terms of the percent in the domain sizes. The number $n$ varies over 2.5\%, 3,75\%, 5.00\%, 6.25\%, 7.50\%, 8.75\% and 10\% of the domain size. The number $b$ varies over 0.125\%, 0.25\%, 0.625\%, and 1.25\% of the domain size. We also varied the noise level $\sigma$ over $0.1, 0.2, 0.4, 0.6, 0.8, 1$ to emulate different signal-to-noise (SNR) cases. In the case of $\sigma = 1$, the noise level is equal to the maximum signal intensity.  Combining the three parameter values, we have 168 different simulation scenarios, and each scenario is run with 20 replicated runs. Fig. \ref{fig:orig8} illustrates the selected design points obtained with $n=10\%$ and $b=1.25\%$. The design points selected for the first stage are seed locations, selected from the Latin Hypercube Sampling (LHS), and the design points selected for the subsequent stages are more concentrated on jump boundaries and some intensity transitioning areas. 

After all the design points are selected, the noisy observations of the regression function at the design points serve as training data to estimate the regression function. Test locations are randomly sampled from an uniform distribution over the regression domain, excluding the ones overlapped with the training data, and the estimate of the regression function is taken for each of the the test locations, using the procedure in Section \ref{sec:fixed}. The estimates were compared to the corresponding non-noisy regression function values (serving as the ground truth) at the test locations to evaluate the mean square errors. We used two mean square error (MSE) metrics, MSE near jump location curves and MSE in the continuity regions, defined to be
\begin{equation*}
\begin{split}
& \mbox{J-MSE} = \frac{1}{|JB(h)|} \sum_{(x, y) \in JB(h)} (\hat{m}(\V{x}) - m(\V{x}))^2 \\
& \mbox{C-MSE} = \frac{1}{|JB(h)^c|} \sum_{(x,y) \in JB(h)^c} (\hat{m}(\V{x}) - m(\V{x}))^2,
\end{split}
\end{equation*}
where $\hat{m}(\V{x})$ is the jump regression estimate, $JB(h)$ is the set of the test locations whose distance from the closest jump location curve is less than or equal to $h$, and $JB(h)^c$ is the complement of $JB(h)$; $h$ is fixed to be 6, which is about twice of the average distance between two neighboring pixels. 

\subsection{Effect of the tuning parameters, $n$, $b$ and $\sigma^2$}
\begin{figure}[p]
	\centering
	\includegraphics[width=0.85\textwidth]{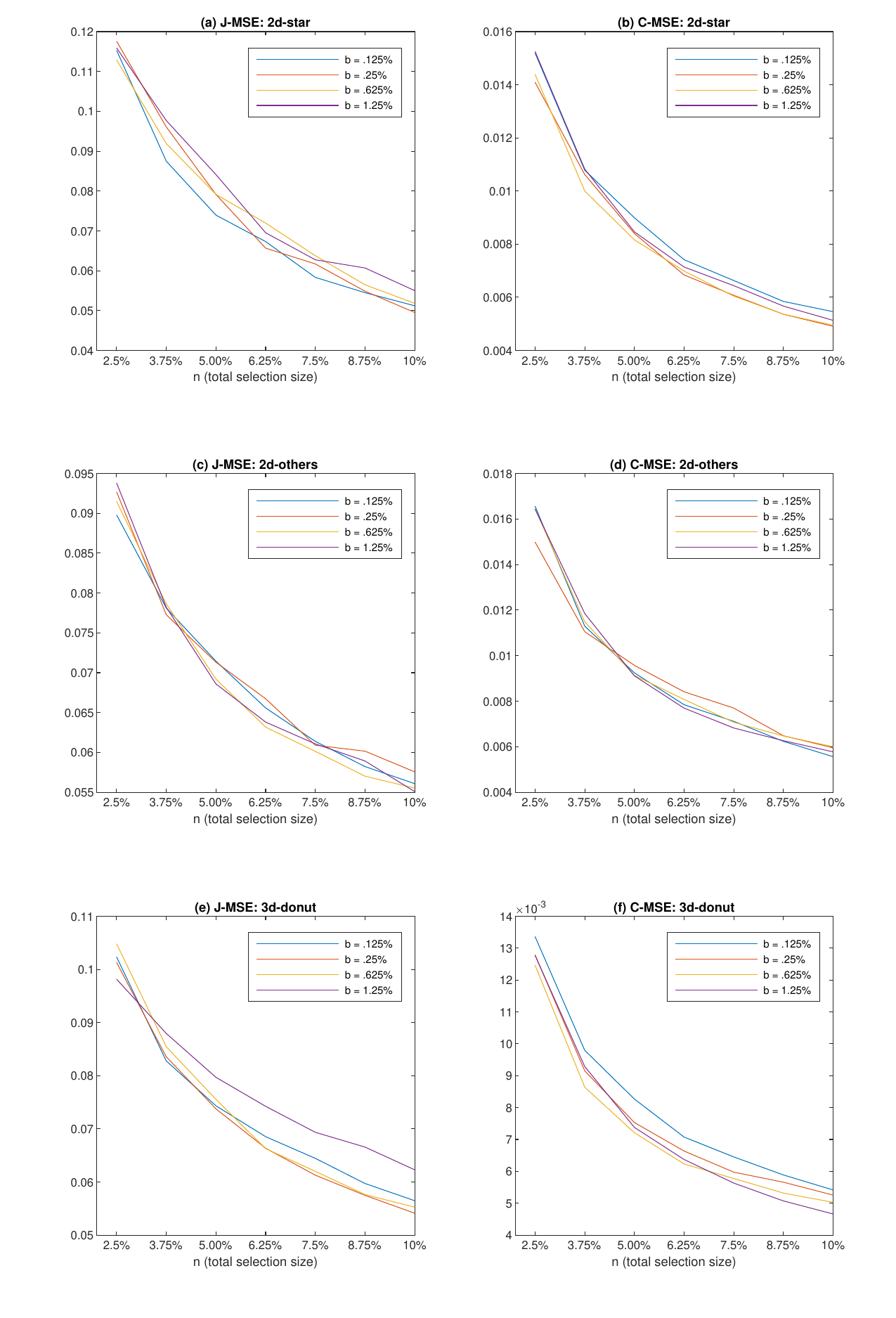}
	\caption{Effect of the total selection size ($n$) and the per-stage selection size ($b$). Here we plot the results for $\sigma=0.4$, because the results for other noise levels follow similar patterns.} \label{fig:orig4}
\end{figure}

We first evaluate how the proposed approach performs for various experimental settings. Fig. \ref{fig:orig4} shows the changes in J-MSE and C-MSE for different settings specified by the total number of the selected design points, denoted by $n$, and the number of the design points selected per stage, denoted by $b$. The per-stage selection size $b$ determines the number of stages for a fixed $n$. According to Fig. \ref{fig:orig4}, the per-stage selection size is not the major factor that affects J-MSE and C-MSE. For the first two test datasets with 2D domains, the per-stage selection size does not make any significant difference in both J-MSE and C-MSE. For the last test dataset with 3D domain, the J-MSE tends to be lower for a smaller $b$ and the C-MSE tends to be lower for a larger $b$. 

Both accuracy measures are more significantly affected by $n$. Based on the results, we would recommend to set $n$ to meet a required level of accuracy and choose a large $b$ for a computational gain. The number of the stages to get $n$ design points is proportional to $n/b$. If $b$ is too small, many stages would be needed, and more frequent computations to update the sampling density function are needed. Therefore, the total computation time would increase. For the remainder of our numerical experiments, we will use $b=1.25\%$, the largest value we tried. We also look at the two performance measures for different noise levels of the observed data. We can see clear downward trends in both J-MSE and C-MSE as the noise level decreases or SNR increases. More details can be found in Appendix B.

\subsection{Comparison to four benchmarks}
We compared the performance of the proposed approach with four benchmark methods. The proposed approach is denoted as JuMp Planner (JMP). The first benchmark is randomly sampling from a uniform density (RAND), the second approach is sampling with Latin Hypercube Sampling (LHS). The third approach is sampling from a density proportional to the weighted residual mean square (WRMS) error of the conventional local linear kernel smoother,
\begin{equation*}
\mbox{WRMS-C}(\V{x}) = \frac{\sum_{\V{x}_i \in \mathcal{N}_n(\V{x})} \left[Y_i - \hat{\alpha} - \hat{\V{\beta}}^T (\V{x}_i - \V{x}) \right]^2  K\left( \frac{\V{x}_i - \V{x}}{h_n(\V{x})}\right)}{\sum_{\V{x}_i \in \mathcal{N}_n(\V{x})} K\left( \frac{\V{x}_i - \V{x}}{h_n(\V{x})}\right)},
\end{equation*}
where $\hat{\alpha}$ and $\hat{\V{\beta}}$ are the optimal solutions of problem \eqref{eq:llocal}. The last benchmark is sampling from a density proportional to the WRMS error of the jump regression model \citep{qiu2004local},
\begin{equation*}
\mbox{WRMS-J}(\V{x}) = \min\{ err^{(1)}(\V{x}), err^{(2)}(\V{x}) \}.
\end{equation*}

\begin{figure}[p]
	\centering
	\includegraphics[width=0.85\textwidth]{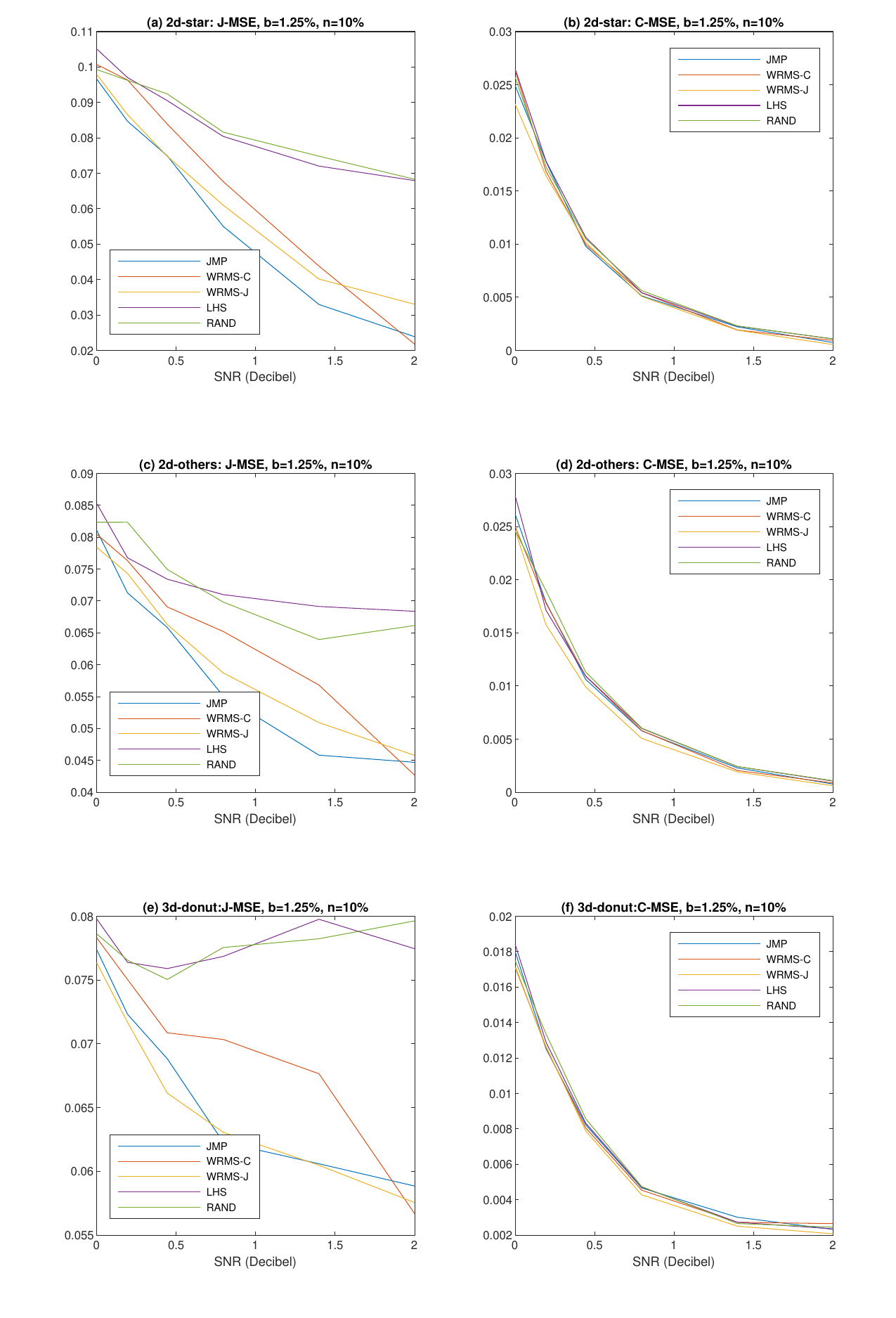}
	\caption{Average Reconstruction Errors over 20 Replications. } \label{fig:error}
\end{figure}

In this comparison, we fixed $b=1.25\%$ and $n=10\%$, because different choices of $b$ and $n$ did not make much difference of the comparison results. Fig. \ref{fig:error} shows SNR versus the averages of the two MSE metrics over 20 replicated simulation runs. From the figure, it can be seen that the C-MSE values computed over the continuity regions do not dependent on the choice of the design selection method. However, the J-MSE values computed near the jump locations differ significantly among different methods. The major findings regarding J-MSE are summarized as below:
\begin{itemize}
	\item Low Noise Case, $\sigma = 0.1$ or SNR = 2: The three error-based methods, JMP, WRMS-J and WRMS-C, significantly outperformed the two random sampling methods, RAND and LHS.
	\item High Noise Case, $\sigma=1$ or SNR = 0 (i.e. maximum intensity of the regression function is equal to $\sigma$): All the compared methods are comparable. It is not very surprising. When the noise level is comparable to the maximum signal intensity, the error-guided methods cannot distinguish signals and noises, so the data sampled would appear pure white noises. In such cases, all three adaptive selection strategies work similarly to the two random sampling methods. 
	\item Medium Noise Cases, $0.1 < \sigma \le 0.8$: The JMP and WRMS-J outperform WRMS-C. This shows the jump regression based approaches are superior to the continuum-based approach around jump regions. The JMP works better than WRMS-J except for 3d-donut example, where both are comparable.  
\end{itemize}

\begin{figure}
	\centering
	\includegraphics[width=0.8\textwidth]{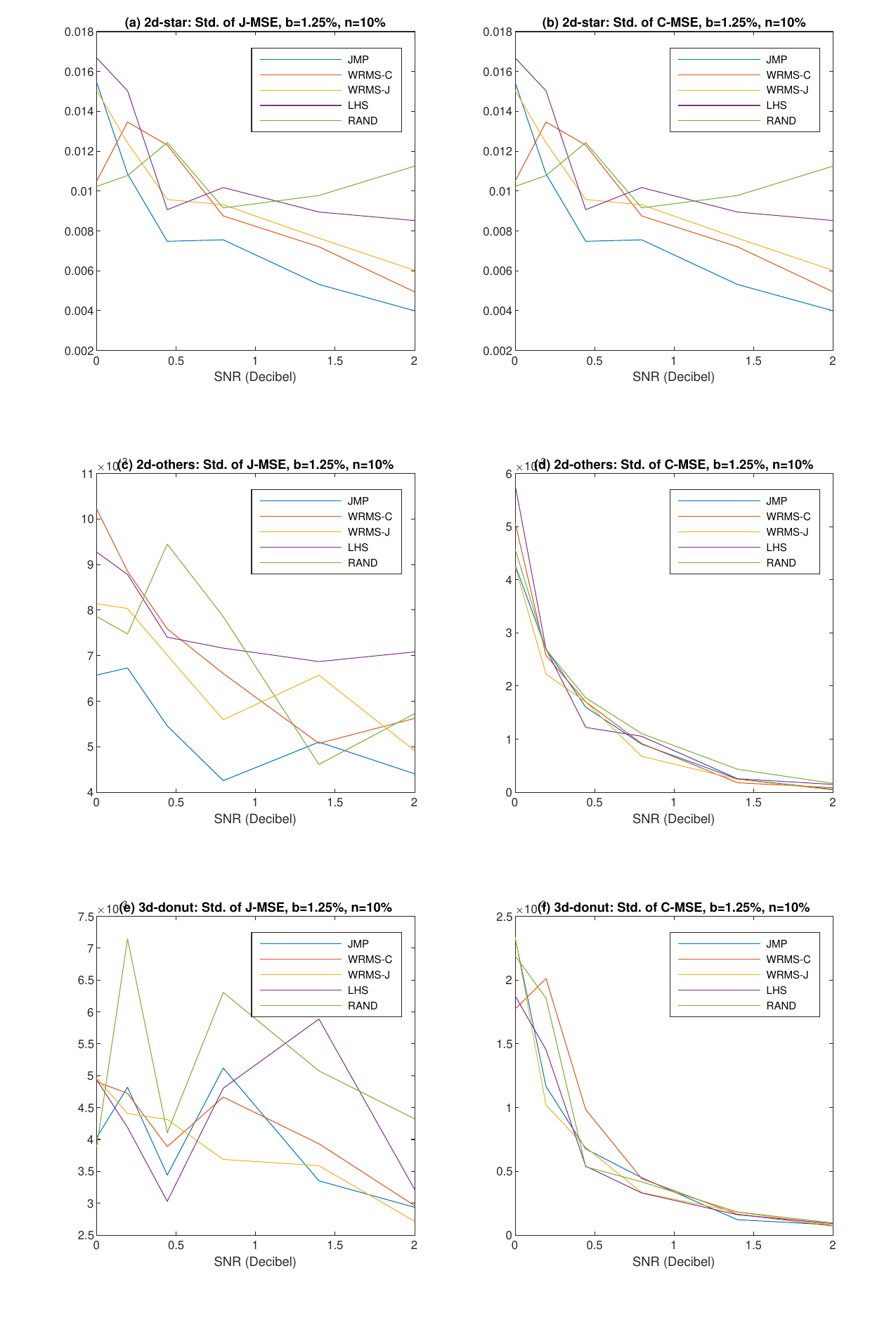}
	\caption{Standard Deviations of Reconstruction Errors over 20 Replications.} \label{fig:error_std}
\end{figure}

We also present the variabilities of the two MSE metrics over 20 replicated simulation runs. Fig. \ref{fig:error_std} in the Online Supplementary Materials (Appendix B) shows the standard deviations of J-MSE and C-MSE values for different noise levels. The overall variabilities of the compared methods increase as $\sigma$ increases or SNR decreases. Among the five methods, the proposed approach has the lowest variability in most cases considered. Thus, a low variability is another advantage of the proposed approach.  

\section{Real Data Study: Adaptive Microscope Imaging For Accelerating the Imaging Speed} \label{sec:app1}
This section presents the application of the proposed approach for compressive and adaptive microscope imaging. 

\subsection{Scientific Background and Significance} \label{sec:app1_1}
Notable improvements in the spatial resolution of scanning transmission electron microscopy (STEM) allow sub-angstrom resolution observations of important material processes to advance materials research. However, the slow speed of the imaging limits the application of the imaging technique to study rapidly changing dynamic processes occurring in nanoscale, such as nanomaterial growth and interactions in response to tumor tissues, rechargeable battery systems, and protein folding. Accelerating the imaging speed would open unprecedented opportunities in studying these important material processes. In conventional STEM imaging, an electron beam is rastered across the specimen under study. The full scan yields a fine-resolution image of the specimen, but it takes a significant amount of time. The imaging time is proportional to the total pixel number, which is very large for a fine-resolution image. One method used to increase the alacrity of imaging is to collect only a targeted, partial set of pixels. In a partial scan, the selection of the pixel locations to be scanned is important to minimize the information loss due to reduction in the number of measurements. Another issue is that the computation time spent to select the good partial scan cannot be very long, because the main purpose of taking a partial scan is to accelerate the imaging speed. If the computation time is very long, it will cancel out the benefit of the shortened physical scan time from the partial scan. Therefore, the computation time cannot surpass the full scan time if one wants to accelerate the total imaging acquisition time. This time requirement makes many Bayesian tree-based sequential design approaches ineligible for this application. In this section, we illustrate the use of the proposed sequential design selection approach to sequentially select the locations of the partial scan. The sequential selection works in multiple stages of partial scans, and the information obtained from the previous stages are exploited to optimize the selection of the next stage.  

\subsection{Application Details}
In this application, the total sample size $n$ is set to be $10\%$ of the pixel counts of the full scan counterpart, considering the accuracy requirement. Based on prior numerical trials, the 10\% partial scan provided a good accuracy for estimating the underlying images. Lowering the sample size would loss many sharp features of the material images, and increasing the sample size would increase the imaging time. In STEM imaging, the time to scan one pixel is referred to as a pixel dwell time, which is about 10 to 40 microseconds. To achieve good quality pixel measurements, 40 microseconds of the pixel dwell time is applied. The total imaging time is approximately the number of pixels to scan multiplying by the pixel dwell time. For example, scanning a $587 \times 484$ imaging area would take $587 \times 484 \times 40 \mu s$, equal to $10.9$ seconds in total. If we only select 10\% of the imaging pixels for a partial scan, then the physical scanning time would be only 1.09 seconds, i.e., 10\% of the full scan time. 

We set the per-stage selection size $b$ to be $1.67\%$, considering the imaging acceleration factor; the choice of $b=1.67\%$ for $n=10\%$ implies that there will be six stages. The number of the stages is denoted by $M$, i.e., $M=6$. As we discussed in Section \ref{sec:app1_1}, selecting design points over multiple stages would require a significant computation time for each stage, because the sampling density is recalculated every stage based on the results from all past stages. The total computation time increases as the number of stages, $M$, increases. Therefore,  the number $M$ in the sequential design selection scheme should be selected carefully, considering the computation time and the accuracy of the image reconstruction with the samples. We performed some initial experiments to choose the appropriate number for $M$. Table \ref{tbl:comp_time} shows the computation times and reconstruction accuracy for different values of $M$ when 10\% of a $587 \times 484$ test image is sub-sampled using the proposed approach:\\
\begin{table}[h]
	\begin{tabular}{|c|c|c|c|c|c|c|c|c|c|}
		\hline
		No. Stages $(M)$ &  M=2    	&   M=3  	&  M=4 	& M=5 	& M=6		& M=7     & M =8   &  M=9 \\
		\hline
		Computation Time (sec.) & 1.5947  &  2.3821 &   2.8685 &   2.8321  &  3.0308  &  3.3091  &  3.7144 & 3.9679  \\ 
		Reconstruction Error & 0.0195  &  0.0195  &  0.0190  &  0.0200  &  0.0198  &  0.0195  &  0.0194  &  0.0203 \\
		\hline
	\end{tabular}
\caption{Computation time and reconstruction error versus the number of stages for the first image in Fig. \ref{fig:mgdata} when n = 10\%.} \label{tbl:comp_time}
\end{table}

The computation time increases as $M$ increases, but the reconstruction error changes very little. We have the similar outcomes for other test images. This implies that it is better to choose a smaller $M$ for a computational gain while maintaining the almost same reconstruction accuracy.  If the number of stages was $M=6$, the total time of the partial scan would include 1.09 seconds of the physical scan time plus 3.0308 seconds of the computation time, which would be 2.5 times shorter than the time for the full scan. The imaging accelerating factor would be 2.5 in such a case. We will use $M=6$ for the remainder of this section, which corresponds to $b=1.67\%$. 

\subsection{Applications to STEM Imaging Under Various Different Conditions}
\begin{figure}[t]
	\includegraphics[width=\textwidth]{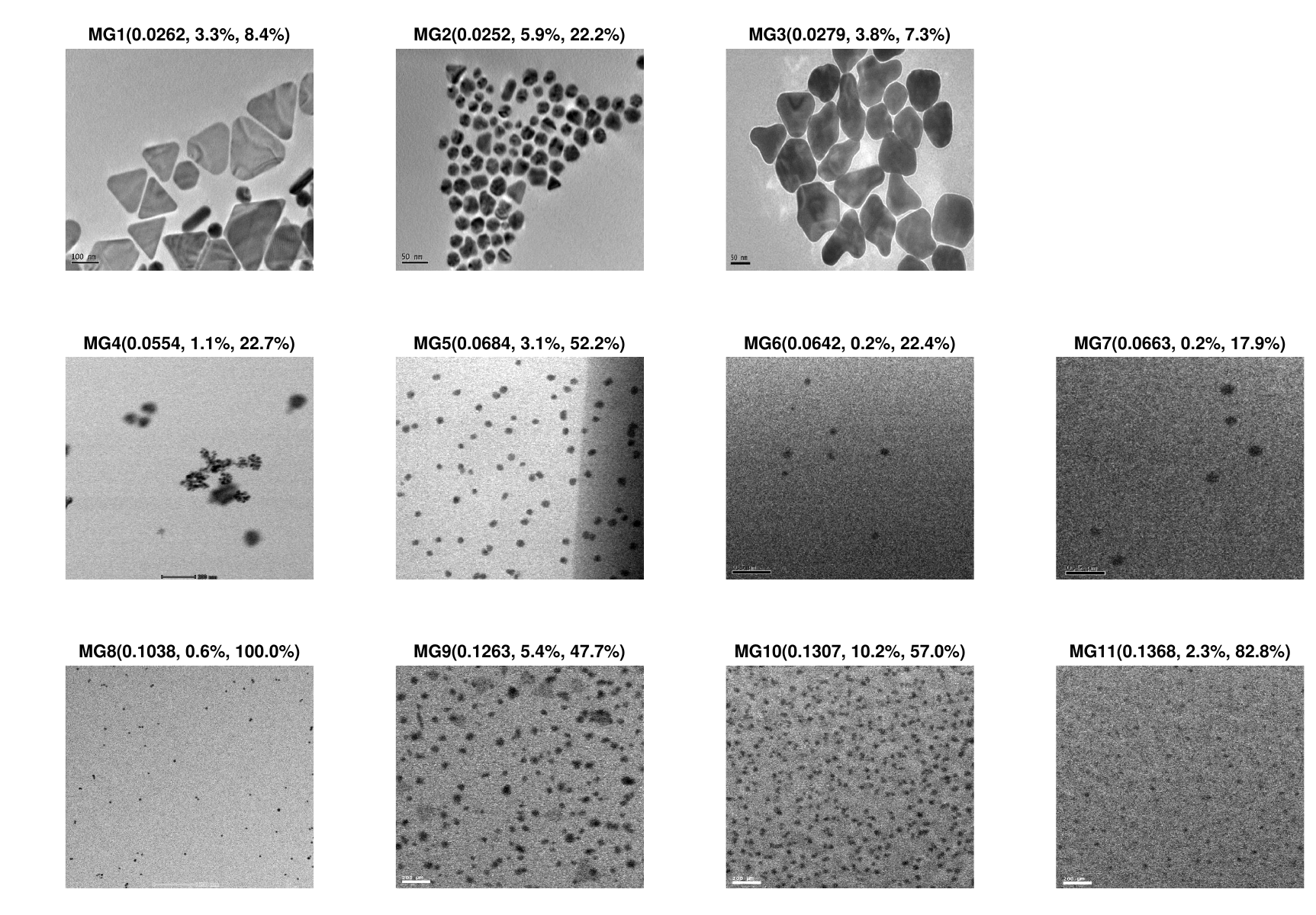}
	\caption{Full raster scanned microscope images. Each image was labeled with image number ($\sigma$, $r$, $s$), where $\sigma$ is the noise standard deviation when the image intensity is normalized so that its maximum is 1, $r$ is the ratio of the foreground boundary pixel number to the total image pixel number, and $s$ is the ratio of the foreground boundary pixel number to the foreground pixel number.} \label{fig:mgdata}
\end{figure}

To quantitatively evaluate our approach compared to the standard STEM imaging techniques, we first obtained complete imaging scans for eleven different specimens (Fig. \ref{fig:mgdata}) to serve as ground truth. These microscope images are characterized by their noise level, ratio of foreground-boundary pixel number to total-image pixel number (FR), and the ratio of foreground-boundary pixel number to foreground pixel number (BFR). The noise levels of the images are estimated. We first took the jump regression estimates of the images for denoising, and the noise variances are estimated by taking the mean squared differences of the estimated regression surfaces and the corresponding original images. The product of FR and BFR quantifies the ratio of the jump location curve pixel number relative to the total pixel number. The eleven images have $587 \times 484$, $587 \times 465$, $611 \times 474$, $592 \times 592$, $472 \times 459$, $1006 \times 1006$, $793 \times 916$, $579 \times 579$, $505 \times 500$, $501 \times 498$, and $502 \times 496$ pixels, respectively. 

In each of the eleven cases, we also achieved the partial scan using our proposed approach. The number selected is equivalent to ten percent of the total raster location number, and the locations in the subset were selected sequentially over six stages. The subset of the raster locations and the corresponding pixel measurements were used to estimate the pixel measurements at the other unselected raster locations. The estimates were compared to the corresponding values from the full raster image, and the two performance metrics, J-MSE and C-MSE, were computed against the noisy full scanned image as $m(\V{x})$. The evaluation of J-MSE requires $JB(h)$, which was estimated. We first applied an image segmentation algorithm to identify the outlines of black regions, and the result of the image segmentation algorithm was manually corrected for a better accuracy. $JB(h)$ is estimated accordingly. Samplings from uniform density, WRMS-C and WRMS-J were used as benchmarks again.

\begin{figure}[t]
	\includegraphics[width=\textwidth]{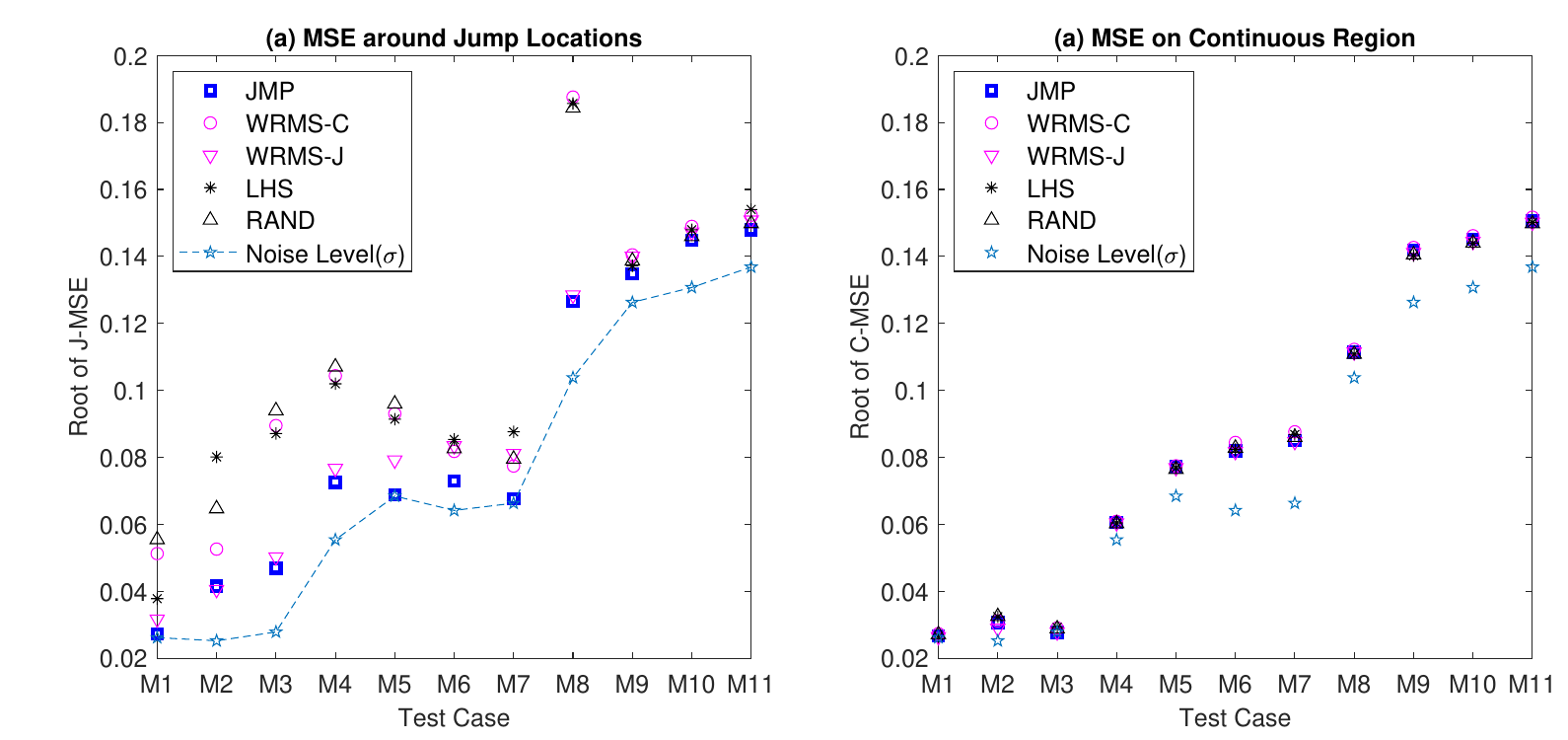}
	\caption{Reconstruction Errors of Different Sampling Methods for 11 Microscope Images.} \label{fig:mgdata_res}
\end{figure}

\begin{figure}[ht!]
	\includegraphics[width=\textwidth]{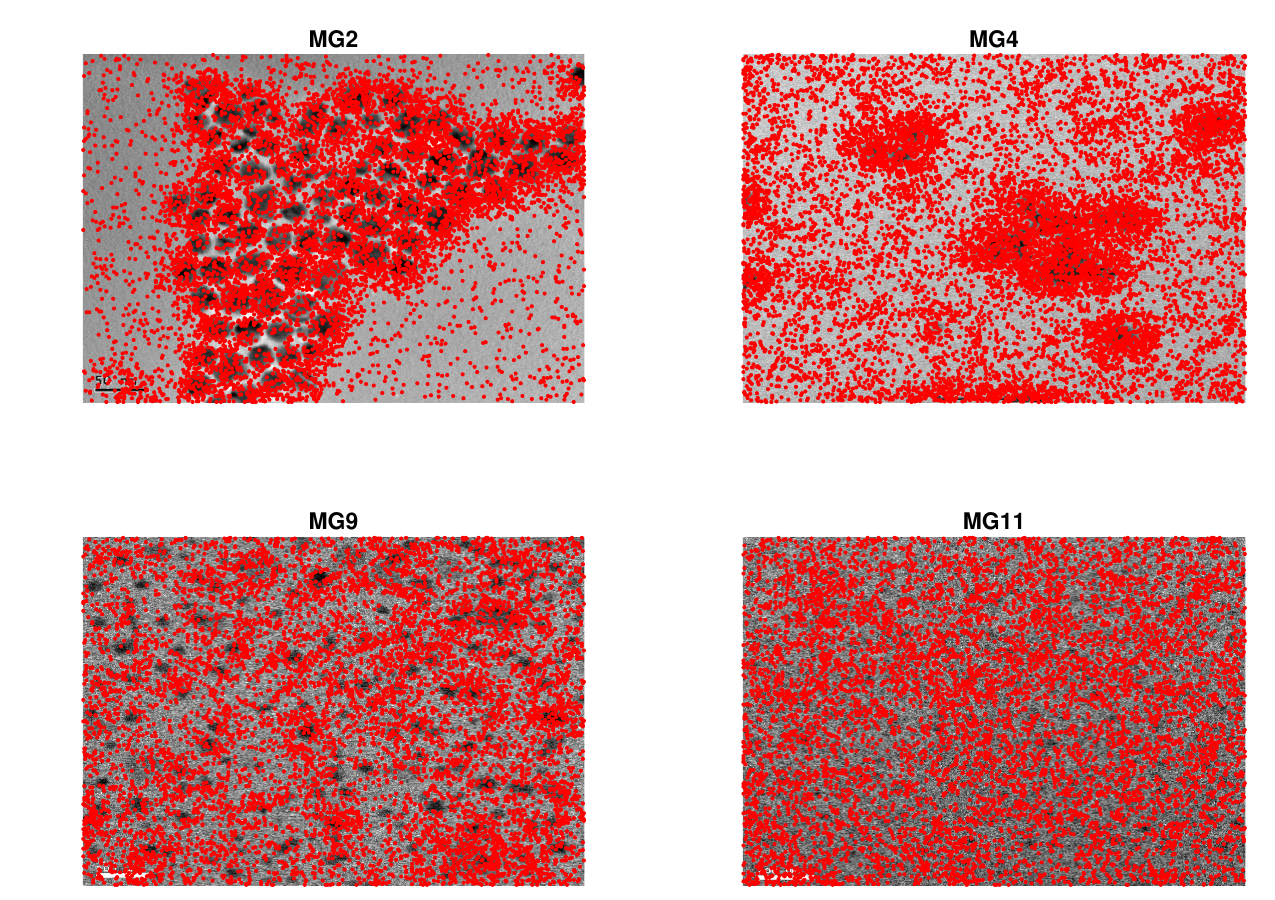}
	\caption{Locations of Partial Scans. Red dots represent the locations for n=3\%. } \label{fig:locs}
\end{figure}

Fig. \ref{fig:mgdata_res} shows the comparison of the related methods in terms of the two performance metrics. Fig. \ref{fig:locs} shows the design point locations selected by the partial scans as red dots. We used $n=3\%$ for the illustration because the red dots are too dense to show the results effectively otherwise. A few key findings are summarized below.
\begin{itemize}
	\item For all test images, the Root C-MSE values for different design selection strategies are comparable and are very close to the noise level. This is consistent with what we found in the simulation study.
	\item MG1 through MG4 (Low Noise and High Ratio of Jump Location Boundary Pixels): Sampling from WRMS-C and the proposed design selection strategy are significantly better than the other methods. This is also consistent with the findings from the simulated studies. 
	\item MG5 through MG7 (Medium Noise): The proposed  design selection approach is better than all the other methods with significant margins, while sampling from WRMS-C is not much better than Random Sampling.
	\item MG8 (High Noise and Many Tiny Foregrounds): Sampling from WRMS-C and the proposed  design selection strategy are significantly better than sampling from WRMS-J and Random Sampling. 
	\item MG9 through MG11 (Very High Noise): All compared methods perform similarly. The proposed strategy is based on the jump detection statistic, which is almost uniform when the noise level is comparable to the jump size $c_{J}$, so the strategy becomes closer to the uniform sampling strategy as shown in Figure \ref{fig:locs} of Appendix C. 
\end{itemize} 

In summary, the proposed approach is very promising in compressive and adaptive imaging for accelerating the image scan speed in STEM unless the level of image noise is comparable to the intensity jumps at edges. 
 
\section{Real Data Study: Experimental Campaign for Predicting Carbon Nanotube Growth} \label{sec:app2}
This section presents the application of the proposed approach to another motivating example of this paper, a problem of optimizing an experimental campaign for predicting the response variable of a chemical experiment under a given experimental condition when the response jumps around certain characteristic boundaries. 

\subsection{Scientific Background and Significance} 
We use a chemical experiment of carbon nanotube growth as a motivating example. Carbon nanotubes are tubes made of carbon atoms with nano-scale diameters. The nanotubes exhibit exceptional tensile strengths and great thermal/electricity conductivity, which are being applied for many practical applications. They are chemically synthesized using a chemical vapor deposition (CVD) process. We are interested in understanding how the reaction conditions of the chemical process affect the carbon nanotube growth. The dependent variable of interest (i.e. the reponse variable) is the resultant amount of carbon nanotube growth under a given reaction condition, and the input variables describe the reaction condition. Among many process parameters describing the reaction condition, the reaction temperature and the composition of chemical reactants greatly affect the growth outcomes, which are the two experimental inputs. The first chemical reactant is C$_2$H$_4$, which is a catalyst to promote the growth reaction. The second reactant is CO$_2$, which suppress the growth reaction. When the concentration ratio of the two chemicals is below a certain threshold, the amount of the carbon nanotube growth is flattened to almost a zero level, but the amount suddenly jumps to a certain level right above the threshold; the observed jump behaviors in the closed-loop carbon nanotube (CNT) growth are a direct result of catalyst phase transition, and the underlying physics is discussed in greater detail in our upcoming publication \citep{carpena2020}. 

Estimating the response surface embedding jumps would require a significant number of experiments if one uses an uniform design such as a space filling design, mainly due to the presence of sharp jumps in the response surface and locating the sharp jumps precisely is possible only when the design points are uniformly dense over the design space unless a non-uniform design is adopted. From a past experimental campaign of the same kind done at AFRL, about 70 design points were selected manually by a human operator, and many of the design points were located in not much useful zero-flat growth regions which yielded a rough estimation of the underlying responses. To make the experimental campaign more efficient, we applied our proposed  design selection strategy to select an experimental design for estimating the response surface for better estimating the response surface with increased fidelity in locating the jump structures. The accurate estimation of the response surface and the embedded jumps would guide practitioners to design their CVD processes for good carbon nanotube yields.  

\subsection{Application Details}
The sequential process is implemented by Air Force Research Lab (AFRL) using a research robot, Autonomous Research System (ARES), that performs CNT growth experiments. The detailed description of the growth experiments can be found in our previous works \citep{nikolaev2016autonomy, rao2012situ, nikolaev2014discovery}. We limit the design space into practical ranges of the two input variables. The practical range for the concentration ratio of the two chemical reactants is from 0 to 6.7 in the log scale or from 1 to 800. The ratio below zero cannot expect any growth, because there is more growth suppressor (CO$_2$) than the catalyst. The ratio 800 is regarded as almost pure catalysts, so the further increase of the ratio would not be more effective. The reaction temperature ranges from 600 to 1100 Celsius. The temperature below 550 is too low to induce the growth of carbon nanotubes, and the temperature above 1200 is difficult to apply given a heat source and the melting temperature of supporting materials. The design space would be $[0, 6.7] \times [600, 1100]$ of the log ratio and temperature. We use the proposed multi-stage sequential design approach for exploring the response surface over the design space. The first stage is the seed experiment, and the experimental design of the first stage is hand-picked by an expert. For running experiments efficiently, five distinct values of the concentration ratio are tried, which are 1, 10, 100, 400 and 680 or 0, 2.3, 4.6, 6.0 and 6.5 in the log scale. For each of the five values, five to seven reaction temperatures are tried, and the reaction temperatures are hand-picked from the temperature range where jumps in growth are expected based on some prior engineering knowledge. In total, 31 design points are selected for the seed stage. In the second stage and thereafter, the design points are chosen by our proposed approach. Given experimental costs, each stage cannot perform too many experiments, so we run 20 experiments per stage. The stages continue until we have a satisfactory outcome. Therefore, the total selection size $n$ is adaptively chosen.   

\begin{figure}[ht!]
	\includegraphics[width=\textwidth]{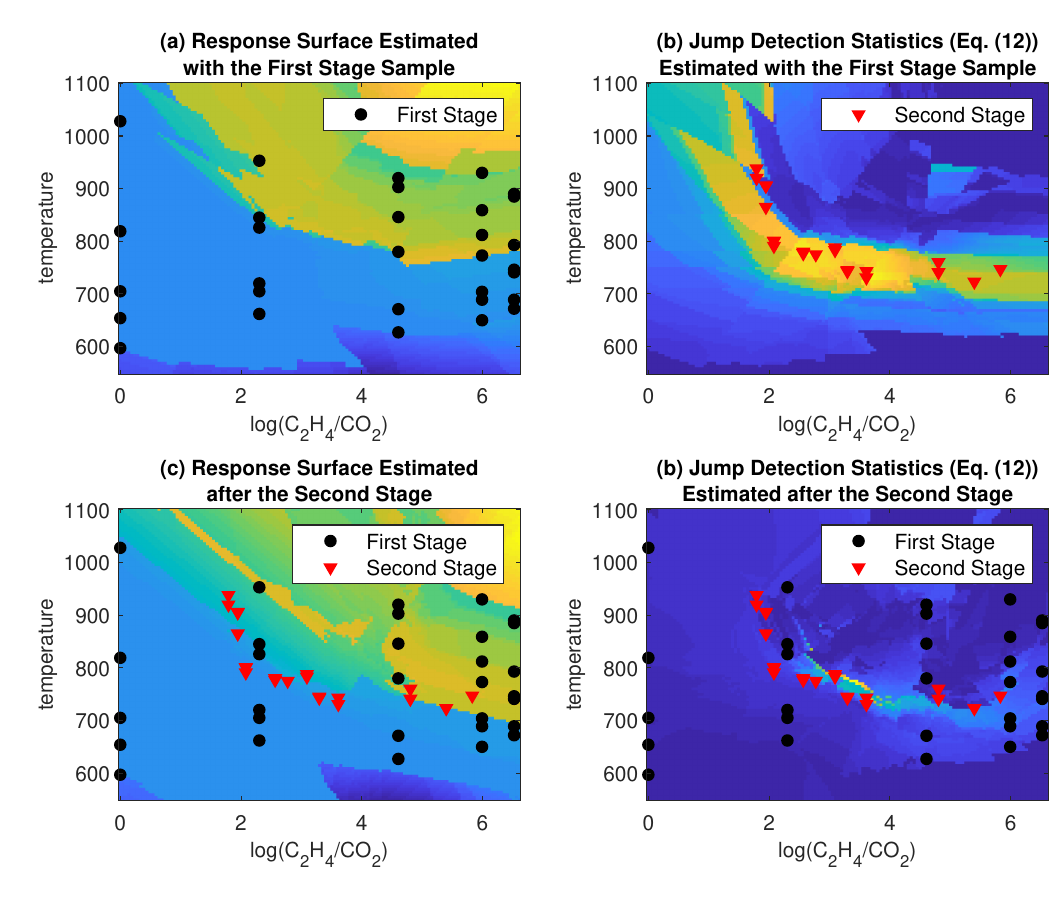}
	\caption{Application of the Proposed Sequential Design Approach. (a) shows the response surface estimate with the first stage seed experiment. (b) shows the jump detection statistic in equation \eqref{eq:jump_stat} estimated with the first stage experiment. In (b), red triangules represent the second stage sample from the sampling density $f_{2|1}$ that is calculated using the jump detection statistic. (c) shows the response surface estimate updated with the second stage experiment. (d) shows the jump detection statistic updated with the second stage experiment. In (d), the jump region is narrowed down to a thin layered region, around which there are many design points already sampled.} \label{fig:ares}
\end{figure}

\subsection{Results}
Fig. \ref{fig:ares}-(a) shows the response surface estimated with the experimental outcomes at the 31 design points of the first stage, and Fig. \ref{fig:ares}-(b) shows the jump detection statistics (equation \eqref{eq:jump_stat}) estimated with the first stage sample. The yellow band with a quite thick bandwidth in the figure is the potential jump region. The thick bandwidth implies that the region of jumps in nanotube growth is not narrowed down, so we need to take more experiments to narrow it down. Based on the statistics and the corresponding sampling density $f_{2|1}$, the second stage samples are taken as shown in Fig. \ref{fig:ares}-(b). The samples are mostly from the yellow band. After the second stage is completed, the response surface is re-estimated as shown in Fig. \ref{fig:ares}-(c), and the corresponding jump detection statistic is estimated as shown in Fig. \ref{fig:ares}-(d). The jump region is narrowed down enough, so we decided to stop the design selection. In total, we took 51 design points, which is a very small number compared to more than several hundred design points necessary to narrow down the jump region following the uniform design of experiments. 

\section{Conclusion} \label{sec:conclude}
We proposed a novel adaptive design strategy (cf., \eqref{eq:conden}) for sequential selection of design points in jump regression analysis. The proposed method originated from our asymptotic error analysis of the jump regression estimate based on the one-sided local linear kernel smoothing, which showed that placing more design points around the jump location curves would give a faster decay of the integrated mean square regression error. Therefore, the proposed sampling function has a large density around the jump location curves. The proposed strategy was applied to two materials science applications, the compressive material imaging problem in which sub-sampled images are used for reconstructing full images and the design selection for accelerating the materials discovery. The outcomes are promising. We have showed the STEM imaging can be accelerated at least ten times, preserving sharp image features, unless the image noise level is comparable to or higher than the image contrast. We also showed from the second example that experimental campaigns for materials discovery in carbon nanotubes can be accelerated, because the design selection can be optimized using our proposed approach. 

\if0\blind
{
	\section*{Acknowledgment} 
We acknowledge support for this work from the AFOSR (FA9550-18-1-0144), NSF (DMS-1914639) and Oak Ridge National Laboratory (4000152630). 
} \fi

\bibliographystyle{agsm} 
\bibliography{jrs}
\newpage
\section*{Online Supplementary Materials}

\subsection*{Appendix A. Proof of Theorem \ref{thm:1}} 
Suppose that $\V{x} \in \mathcal{A}_b$ and it is non-singular in that $\mathcal{N}_n(\V{x})$ does intersect only with $\mathcal{A}_b$ and one another sub-region, say $\mathcal{A}_{b'}$. The local linear kernel estimate $\hat{m}_{(0)}(\V{x})$ can be expressed as 
\begin{equation*}
	\hat{m}_{(0)}(\V{x}) = \sum_{\V{x}_i \in \mathcal{N}_n(\V{x})} \omega(\V{x},\V{x}_i) Y_i,
\end{equation*}
for a conditional second order kernel $\omega$ that satisfies the following conditions:
\begin{equation*}
	\sum_{\V{x}_i \in \mathcal{N}_n(\V{x})} \omega(\V{x},\V{x}_i) = 1 \mbox{ and } \sum_{\V{x}_i \in \mathcal{N}_n(\V{x})} \omega(\V{x},\V{x}_i) (\V{x}_i - \V{x}) = 0.
\end{equation*}
We use \citet[Theorem 2.1]{ruppert1994multivariate} to get the variance of the estimate to be
\begin{equation*}
	\begin{split}
		Var[\hat{m}_{(0)}(\V{x})] = \frac{\sigma^2}{nh_n^p}  R(K) / f(\V{x}) (1+o_P(1)),
	\end{split}
\end{equation*}
where $R(K) = \int K^2(\V{u})d\V{u}$. Since we choose the spatial varying bandwidth $h_n(\V{x}) \propto n^{-1/p} f(\V{x})^{-1/p}$, 
the variance is asymptotically a constant since
\begin{equation*}
	\begin{split}
		Var[\hat{m}_{(0)}(\V{x})] = \kappa_1 \sigma^2 (1+o_P(1)),
	\end{split}
\end{equation*}
where $\kappa_1$ is a fixed constant.
The expectation of the estimate is 
\begin{equation*}
	E[\hat{m}_{(0)}(\V{x})] = \sum_{\V{x}_i \in \mathcal{N}_n(\V{x})} \omega(\V{x},\V{x}_i) m(\V{x}_i).
\end{equation*}
To further analyze the term, let $Q^{(b)}_n = \mathcal{N}_n(\V{x}) \cap \mathcal{A}_{b}$ and $Q^{(b')}_n = \mathcal{N}_n(\V{x}) \cap \mathcal{A}_{b'}$. The expectation can be split accordingly as follows: 
\begin{equation} \label{eq:exp1}
	\begin{split}
		E[\hat{m}_{(0)}(\V{x})] &= \sum_{\V{x}_i \in Q^{(b)}_n} \omega(\V{x},\V{x}_i) m(\V{x}_i) + \sum_{\V{x}_i \in Q^{(b')}_n} \omega(\V{x},\V{x}_i) m(\V{x}_i) \\
		&= \sum_{\V{x}_i \in Q^{(b)}_n} \omega(\V{x},\V{x}_i) g_b(\V{x}_i) + \sum_{\V{x}_i \in Q^{(b')}_n} \omega(\V{x},\V{x}_i) g_{b'}(\V{x}_i)
	\end{split}
\end{equation}
Let $\V{x}_J$ denote a boundary point in $\partial\mathcal{A}_b \cap \partial\mathcal{A}_{b'}$ which is closest from $\V{x}$, and let $c_{J} = g_{b'}(\V{x}_J) - g_b(\V{x}_J)$ denote the intensity jump at the boundary location. We take the second order Taylor expansion of $g_b(\V{x}_i)$ at $\V{x}_J$ as
\begin{equation*}
	g_b(\V{x}_i) = g_{b}(\V{x}_J) + \V{d}_{J,b}^T (\V{x}_i-\V{x}_J)  + \frac{1}{2}(\V{x}_i-\V{x}_J)^T \M{Q}_{J,b} (\V{x}_i - \V{x}_J) + o_P(h_n^2),
\end{equation*}
where $\V{d}_{J,b}$ is the first order partial derivative of $g_b$ at $\V{x}_J$, $\M{H}_{J,b}$ is the Hessian matrix at $\V{x}_J$, and the remainder term is bounded by $h_n$. 
With the Taylor expansion, the expectation \eqref{eq:exp1} can be written as
\begin{align*}
	& E[\hat{m}_{(0)}(\V{x})] \\
	&= \sum_{\V{x}_i \in Q^{(b)}_n} \omega(\V{x},\V{x}_i) \{ g_{b}(\V{x}_J) + \V{d}_{J,b}^T (\V{x}_i-\V{x}_J)  + \frac{1}{2}(\V{x}_i-\V{x}_J)^T \M{H}_{J,b} (\V{x}_i - \V{x}_J) \}\\
	& + \sum_{\V{x}_i \in Q^{(b')}_n} \omega(\V{x},\V{x}_i) \{ g_{b'}(\V{x}_J) + \V{d}_{J,b'}^T (\V{x}_i-\V{x}_J)  + \frac{1}{2}(\V{x}_i-\V{x}_J)^T \M{H}_{J,b'} (\V{x}_i - \V{x}_J)  \} \\
	&  +o_P(h_n^2)\\
	&= \sum_{\V{x}_i \in Q^{(b)}_n} \omega(\V{x},\V{x}_i) \{ g_{b}(\V{x}_J) + \V{d}_{J,b}^T (\V{x}_i-\V{x}_J)  + \frac{1}{2}(\V{x}_i-\V{x}_J)^T \M{H}_{J,b} (\V{x}_i - \V{x}_J) \\
	& + \sum_{\V{x}_i \in Q^{(b')}_n} \omega(\V{x},\V{x}_i) \{ c_{J} + g_{b}(\V{x}_J) + \V{d}_{J,b'}^T (\V{x}_i-\V{x}_J)  + \frac{1}{2}(\V{x}_i-\V{x}_J)^T \M{H}_{J,b'} (\V{x}_i - \V{x}_J)  \} \\
	& + o_P(h_n^2)\\
	&= \sum_{\V{x}_i \in \mathcal{N}_n(\V{x})} \omega(\V{x},\V{x}_i) \{ g_{b}(\V{x}_J) + \V{d}_{J,b}^T (\V{x}_i-\V{x}_J)  + \frac{1}{2}(\V{x}_i-\V{x}_J)^T \M{H}_{J,b} (\V{x}_i - \V{x}_J)  \} \\
	& + \sum_{\V{x}_i \in Q^{(b')}_n} \omega(\V{x},\V{x}_i) \{ c_{J} + \V{\delta}_{J}^T (\V{x}_i-\V{x}_J)  + \frac{1}{2}(\V{x}_i-\V{x}_J)^T \M{\Delta}_J  (\V{x}_i - \V{x}_J) \} + o_P(h_n^2),
\end{align*} 
where $\V{\delta}_{J} = \V{d}_{J,b'}-\V{d}_{J,b}$, and $\M{\Delta}_J = \M{H}_{J,b'}-\M{H}_{J,b}$. With the Taylor expansion of $m(\V{x})$ at $\V{x}_J$, 
\begin{equation} \label{eq:a}
	\begin{split}
		m(\V{x}) = g_b(\V{x}_J) +  \V{d}_{J,b}^T (\V{x}-\V{x}_J)  + \frac{1}{2}(\V{x}-\V{x}_J)^T \M{H}_{J,b} (\V{x} - \V{x}_J) + o_P(h_n^2), 
	\end{split}
\end{equation}
the bias of $\hat{m}_{(0)}(\V{x})$ is 
\begin{equation*}
	\begin{split}
		& E[\hat{m}_{(0)}(\V{x})] - m(\V{x}) \\
		& \quad = \sum_{\V{x}_i \in \mathcal{N}_n(\V{x})} \omega(\V{x},\V{x}_i) \{ (\V{d}_{J,b} - \M{H}_{J,b}\V{x}_J + 2\M{H}_{J,b}\V{x})^T (\V{x}_i-\V{x})  + \frac{1}{2}(\V{x}_i-\V{x})^T \M{H}_{J,b} (\V{x}_i-\V{x}) \}  \\
		& \qquad + \sum_{\V{x}_i \in Q^{(2)}_n} \omega(\V{x},\V{x}_i) \{ c_{J} + \V{\delta}_{J}^T (\V{x}_i-\V{x}_J)  + \frac{1}{2}(\V{x}_i-\V{x}_J)^T\M{\Delta}_J  (\V{x}_i - \V{x}_J) \} + o_P(h_n^2) \\
		&\quad = \sum_{\V{x}_i \in \mathcal{N}_n(\V{x})} \omega(\V{x},\V{x}_i) \frac{1}{2}(\V{x}_i-\V{x})^T \M{H}_{J,b} (\V{x}_i-\V{x}) \\
		& \qquad + \sum_{\V{x}_i \in Q^{(2)}_n} \omega(\V{x},\V{x}_i) \{ c_{J} + (\V{\delta}_{J} + \M{\Delta}_J\V{n}_J)^T  (\V{x}_i-\V{x})  + \frac{1}{2}(\V{x}_i-\V{x})^T\M{\Delta}_J (\V{x}_i - \V{x})\}\\
		& \qquad + \frac{1}{2}(\M{\Delta}_J\V{n}_J + 2\V{\delta}_{J})^T \V{n}_J  \sum_{\V{x}_i \in Q^{(2)}_n} \omega(\V{x},\V{x}_i)  \\
		&\quad = \sum_{\V{x}_i \in \mathcal{N}_n(\V{x})} \omega(\V{x},\V{x}_i) \frac{1}{2}(\V{x}_i-\V{x})^T \M{H}_{J,b} (\V{x}_i-\V{x}) \\
		& \qquad + \sum_{\V{x}_i \in Q^{(2)}_n} \omega(\V{x},\V{x}_i) \left\{ c_{J} + c\V{\delta}_{J}^T  (\V{x}_i-\V{x})  + \frac{1}{2}(\V{x}_i-\V{x})^T\M{\Delta}_J (\V{x}_i - \V{x}) \right\}\\
	\end{split}
\end{equation*} 
where $\V{n}_J = \V{x} - \V{x}_J$, and $\M{H}_{\tau}\V{n}_J + \V{d}_{\tau} = c \V{d}_{\tau}$ for a constant $c$. In the last equation, the first term is the same as the bias of the estimate in cases when there is no jump around $\V{x}$, and the second term is the contribution of the nearby jump to the bias. Using the result for the local linear kernel estimation \citep{ruppert1994multivariate}, the first term is
\begin{equation*}
	\sum_{\V{x}_i \in \mathcal{N}_n(\V{x})} \omega(\V{x},\V{x}_i) \frac{1}{2}(\V{x}_i-\V{x})^T \M{H}_{J,b} (\V{x}_i-\V{x}) = \frac{1}{2}\mu_2(K)\left(h_n^2 \sum_{j=1}^d \frac{\partial^2 g(\V{x})}{\partial x_j^2}\right) + o_P(h_n^2),
\end{equation*}
where $\mu_2(K)$ is a kernel-dependent constant with $\mu_2(K)\M{I} = \int \V{u}\V{u}^TK(\V{u})d\V{u}$. Since $g$ is smooth with a bounded second derivative, 
\begin{equation} \label{eq:dev}
	\sum_{\V{x}_i \in \mathcal{N}_n(\V{x})} \omega(\V{x},\V{x}_i) \frac{1}{2}(\V{x}_i-\V{x})^T \M{H}_{J,b} (\V{x}_i-\V{x}) = o_P(h_n^2) = o_P\left( n^{-2/p} f(x)^{-2/p}\right).
\end{equation}
Using \citet[Theorem 2.1]{mack1979multivariate},
\begin{equation*} 
	\begin{split}
		& \sum_{\V{x}_i \in Q^{(b')}_n} c_J \omega(\V{x},\V{x}_i) = \left(\frac{1}{f(x)} + o_P(1)\right) n^{-1} \sum_{\V{x}_i \in \mathcal{Q}^{(b')}_n} K\left( \frac{\V{x}_i-\V{x}}{h_n} \right) c_{J}\\
		& \sum_{\V{x}_i \in Q^{(b')}_n} \omega(\V{x},\V{x}_i) \V{\delta}_{J}^T  (\V{x}_i-\V{x}) = \left(\frac{1}{f(x)} + o_P(1)\right) n^{-1} \sum_{\V{x}_i \in \mathcal{Q}^{(b')}_n} K\left( \frac{\V{x}_i-\V{x}}{h_n} \right) \V{\delta}_{J}^T  (\V{x}_i-\V{x}) \\
		& \sum_{\V{x}_i \in Q^{(b')}_n} \omega(\V{x},\V{x}_i) (\V{x}_i-\V{x})^T\M{\Delta}_J (\V{x}_i - \V{x}) \\
		& \qquad = \left(\frac{1}{f(x)} + o_P(1)\right) n^{-1} \sum_{\V{x}_i \in \mathcal{Q}^{(b')}_n} K\left( \frac{\V{x}_i-\V{x}}{h_n} \right) (\V{x}_i-\V{x})^T\M{\Delta}_J (\V{x}_i - \V{x})\\
	\end{split}
\end{equation*}
and the second term in the bias expression is
\begin{equation*}
	\begin{split}
		& \sum_{\V{x}_i \in Q^{(2)}_n} \omega(\V{x},\V{x}_i) \left\{ c_{J} + c\V{\delta}_{J}^T  (\V{x}_i-\V{x})  + \frac{1}{2}(\V{x}_i-\V{x})^T\M{\Delta}_J (\V{x}_i - \V{x}) \right\} \\
		& \qquad = \left(\frac{1}{f(x)} + o_P(1)\right) n^{-1} \sum_{\V{x}_i \in \mathcal{Q}^{(b')}_n} K\left( \frac{\V{x}_i-\V{x}}{h_n} \right)   \left\{ c_{J} + c\V{\delta}_{J}^T  (\V{x}_i-\V{x})  + \frac{1}{2}(\V{x}_i-\V{x})^T\M{\Delta}_J (\V{x}_i - \V{x}) \right\}
	\end{split}
\end{equation*}
where
\begin{equation*}
	\begin{split}
		& n^{-1} \sum_{\V{x}_i \in \mathcal{Q}^{(b')}_n} K\left( \frac{\V{x}_i-\V{x}}{h_n} \right) = \int_{\mathcal{Q}^{(b')}} K(\V{u}) f(\V{x}+h_n\V{u}) d\V{u} + o_P(1) \\
		& \qquad =  \int_{\mathcal{Q}^{(b')}} K(\V{u}) \{f(\V{x}) + h_n D_f(\V{x})^T \V{u} + o(h_n)\} d\V{u} + o_P(1) \\
		& \qquad =   f(\V{x}) \int_{\mathcal{Q}^{(b')}} K(\V{u}) d\V{u} + h_n D_f(\V{x})^T \int_{\mathcal{Q}^{(b')}} \V{u} K(\V{u}) d\V{u} + o_P(1) \\
		& \qquad =   f(\V{x}) \int_{\mathcal{Q}^{(b')}} K(\V{u}) d\V{u} + o_P(1),
	\end{split}
\end{equation*}
\begin{equation*}
	\begin{split}
		& n^{-1} \sum_{\V{x}_i \in \mathcal{Q}^{(b')}_n} K\left( \frac{\V{x}_i-\V{x}}{h_n} \right) \V{\delta}_J^T (\V{x}_i - \V{x})= \int_{\mathcal{Q}^{(b')}} K(\V{u}) h_n\V{u} f(\V{x}+h_n\V{u}) d\V{u} + o_P(h_n) \\
		& \qquad =  \int_{\mathcal{Q}^{(b')}} K(\V{u})  h_n \V{\delta}_J^T\V{u}  \{f(\V{x}) + h_n D_f(\V{x})^T \V{u} + o(h_n)\} d\V{u} + o_P(h_n) \\
		& \qquad =   h_n f(\V{x}) \V{\delta}_J^T\int_{\mathcal{Q}^{(b')}}  \V{u} K(\V{u}) d\V{u} + h_n^2 \V{\delta}_J^T \int_{\mathcal{Q}^{(b')}} \V{u}\V{u}^T K(\V{u}) d\V{u} \cdot D_f(\V{x})  +  o_P(h_n) \\
		& \qquad =   h_n f(\V{x}) \V{\delta}_J^T\int_{\mathcal{Q}^{(b')}}  \V{u} K(\V{u}) d\V{u} + o_P(h_n) \mbox{, and} 
	\end{split}
\end{equation*}
\begin{equation*}
	\begin{split}
		& n^{-1} \sum_{\V{x}_i \in \mathcal{Q}^{(b')}_n} K\left( \frac{\V{x}_i-\V{x}}{h_n} \right) (\V{x}_i - \V{x})^T \M{\Delta}_J  (\V{x}_i - \V{x}) = \int_{\mathcal{Q}^{(b')}} K\left( \frac{\V{x}_i-\V{x}}{h_n} \right) (\V{x}_i - \V{x})^T \M{\Delta}_J (\V{x}_i - \V{x}) f(\V{x}_i) d\V{x}_i \\
		& \qquad =  \int_{\mathcal{Q}^{(b')}} K(\V{u}) h_n^2 \V{u}^T\M{\Delta}_J\V{u} f(\V{x}+h_n\V{u}) d\V{u} + o_P(h_n^2) \\
		& \qquad =  \int_{\mathcal{Q}^{(b')}} K(\V{u}) h_n^2 \V{u}^T\M{\Delta}_J\V{u} \{f(\V{x}) + h_n D_f(\V{x})^T \V{u} + o_P(h_n)\} d\V{u} + o_P(h_n^2) \\
		& \qquad = h_n^2 f(\V{x})\int_{\mathcal{Q}^{(b')}} \V{u}^T\M{\Delta}_J\V{u} K(\V{u}) d\V{u} + h_n^3 \int_{\mathcal{Q}^{(b')}} K(\V{u}) \V{u}^T\M{\Delta}_J\V{u} \V{u}^T d\V{u} D_f(\V{x}) + o_P(h_n^2) \\
		& \qquad = h_n^2 f(\V{x}) tr\left(\M{\Delta}_J \int_{\mathcal{Q}^{(b')}} \V{u}\V{u}^T K(\V{u}) d\V{u} \right) + o_P(h_n^2).
	\end{split} 
\end{equation*}
Therefore, the second term in the bias expression is asymptotically to be
\begin{equation} \label{eq:dev1}
	\begin{split}
		c_{J} \int_{\mathcal{Q}^{(b')}} K(\V{u}) d\V{u} + h_n \V{\delta}_J^T\int_{\mathcal{Q}^{(b')}}  \V{u} K(\V{u}) d\V{u} + h_n^2 tr\left(\M{\Delta}_J \int_{\mathcal{Q}^{(b')}} \V{u}\V{u}^T K(\V{u}) d\V{u} \right) + o_P(h_n^2),
	\end{split}
\end{equation}
where $Q^{(b')}$ is the part of the support of $K$ that corresponds to $Q^{(b')}_n$.

Based on the results of \eqref{eq:dev} and \eqref{eq:dev1}, the bias can be described as
\begin{equation} \label{eq:bias}
	\begin{split}
		E[\hat{m}_{(0)}(\V{x})] - m(\V{x}) =& o_P\left( \frac{1}{n^{2/p} f(x)^{2/p}}\right) \\
		&+ c_{J} \int_{\mathcal{Q}^{(b')}} K(\V{u}) d\V{u} \\
		& + h_n \V{\delta}_J^T\int_{\mathcal{Q}^{(b')}}  \V{u} K(\V{u}) d\V{u} \\
		&+ h_n^2 tr\left(\M{\Delta}_J \int_{\mathcal{Q}^{(b')}} \V{u}\V{u}^T K(\V{u}) d\V{u} \right)\\
		=& o_P\left( \frac{1}{n^{2/p} f(x)^{2/p}}\right) + (c_{J} + o_P(1))\int_{\mathcal{Q}^{(b')}} K(\V{u}) d\V{u}.
	\end{split}
\end{equation}

\section*{Appendix B. J-MSE and C-MSE for different noise levels and different values of $n$}
\begin{figure}[p]
	\centering
	\includegraphics[width=0.85\textwidth]{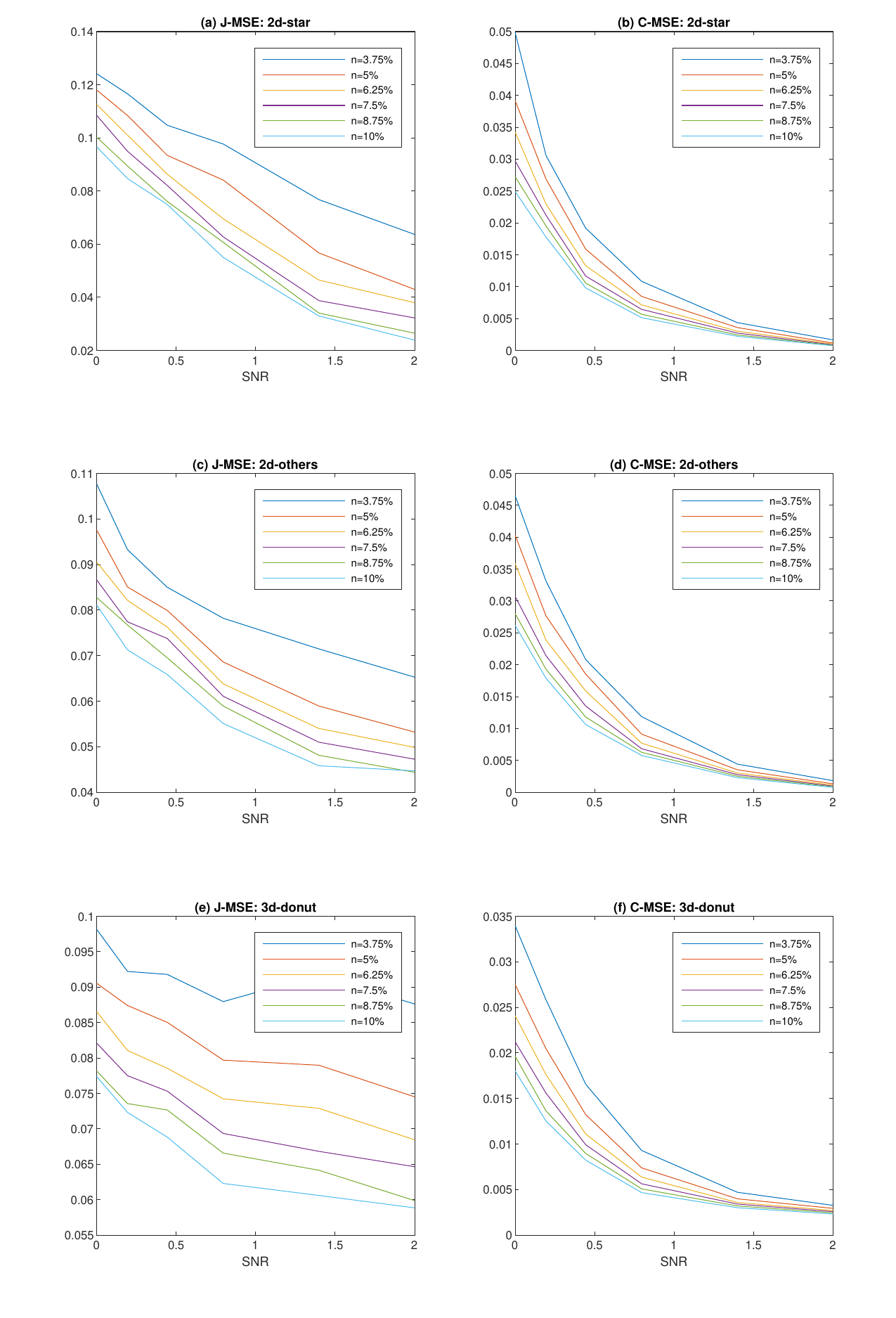}
	\caption{Effect of Noise Level $\sigma$. Here we plot the results for $b=1.25\%$, because the results for other values of $b$ follow similar patterns.} \label{fig:orig5}
\end{figure}

Fig. \ref{fig:orig5} shows J-MSE and C-MSE for different noise levels and different values of $n$. We can see clear downward trends in both J-MSE and C-MSE as the noise level decreases or SNR increases. The C-MSE increases quadratically in $\sigma$, and the J-MSE increases linearly, which may be because the estimation error due to jumps dominates the estimation error due to noise around the jump boundaries.  

\newpage

\subsection*{Appendix C. Numerical illustrations of adaptive STEM scan results discussed in Section \ref{sec:app1}}

\subsection*{Appendix D. Variance of Reconstruction Errors for Synthetic Examples in Section 4}

\end{document}